\documentclass[10pt,twocolumn,letterpaper]{article}
\usepackage{cvpr}   
\definecolor{cvprblue}{rgb}{0.21,0.49,0.74}
\usepackage[pagebackref,breaklinks,colorlinks,allcolors=cvprblue]{hyperref}

\usepackage{graphicx}
\usepackage{amsmath}
\usepackage{amssymb}
\usepackage{booktabs}
\usepackage{float}
\usepackage{comment}
\usepackage{multirow}
\usepackage{makecell}
\usepackage{enumitem}

\usepackage[capitalize]{cleveref}
\crefname{section}{Sec.}{Secs.}
\Crefname{section}{Section}{Sections}
\Crefname{table}{Table}{Tables}
\crefname{table}{Tab.}{Tabs.}

\def\paperID{17477} % *** Enter the Paper ID here
\def\confName{CVPR}
\def\confYear{2025}

\begin{document}

\title{\LARGE \bf 
DGIQA: Depth-guided Feature Attention and Refinement for\\ Generalizable Image Quality Assessment
\vspace{-5mm}
}

\author{Vaishnav Ramesh, Junliang Liu, Haining Wang, and Md Jahidul Islam\\
{\small RoboPI laboratory, Department of ECE, University of Florida}\\
{\small \underline{Model and code:} \url{https://github.com/uf-robopi/DGIQA}}
\thanks{Correspondence: \tt\small vaishnavramesh@ufl.edu}%
}
\maketitle

%%%%%%%%% ABSTRACT
\begin{abstract}
 A long-held challenge in no-reference image quality assessment (NR-IQA) learning from human subjective perception is the lack of objective generalization to unseen natural distortions. To address this, we integrate a novel \textbf{D}epth-\textbf{G}uided cross-attention and refinement (Depth-CAR) mechanism, which distills scene depth and spatial features into a structure-aware representation for improved NR-IQA. This brings in the knowledge of object saliency and relative contrast of the scene for more discriminative feature learning. Additionally, we introduce the idea of TCB (Transformer-CNN Bridge) to fuse high-level global contextual dependencies from a transformer backbone with local spatial features captured by a set of hierarchical CNN (convolutional neural network) layers.  We implement TCB and Depth-CAR as multimodal attention-based projection functions to select the most informative features, which also improve training time and inference efficiency. Experimental results demonstrate that our proposed \textbf{DGIQA} model achieves state-of-the-art (SOTA) performance on both synthetic and authentic benchmark datasets. More importantly, DGIQA outperforms SOTA models on cross-dataset evaluations as well as in assessing natural image distortions such as low-light effects, hazy conditions, and lens flares. 
\end{abstract}

\vspace{-4mm}
\section{Introduction}
\vspace{-1mm}
Image Quality Assessment (IQA) methods provide a means to evaluate the perceptual and statistical \textit{qualities} of images. Traditional full-reference and reduced-reference IQA metrics require an undistorted, high-quality reference image to measure degradation~\cite{wang2006modern}, which is generally unavailable in practice. Therefore, Non-Reference IQA (NR-IQA) methods are essential in computer vision and imaging~\cite{wang2002image,liang2021swinir,wang2021real}, enabling robust prediction of image quality solely based on its \textit{signal} content and statistics~\cite{zhai2020perceptual,li2017statistical,saad2012blind}.

%While classical NR-IQA methods mostly relied on handcrafted features\cite{ahmed2020image} and statistical models\cite{li2017statistical,saad2012blind} to capture distortion characteristics. While these approaches have achieved moderate success, they struggle to generalize across diverse image contents and distortion types.  Afterwords, 
%, making it highly valuable in real-world applications such as quality control system, optimizing the algorithm for image processing\cite{wang2002image}, and image restoration operations\cite{liang2021swinir,wang2021real}.

\begin{figure}[t]
    \centering
    \includegraphics[width=\linewidth]{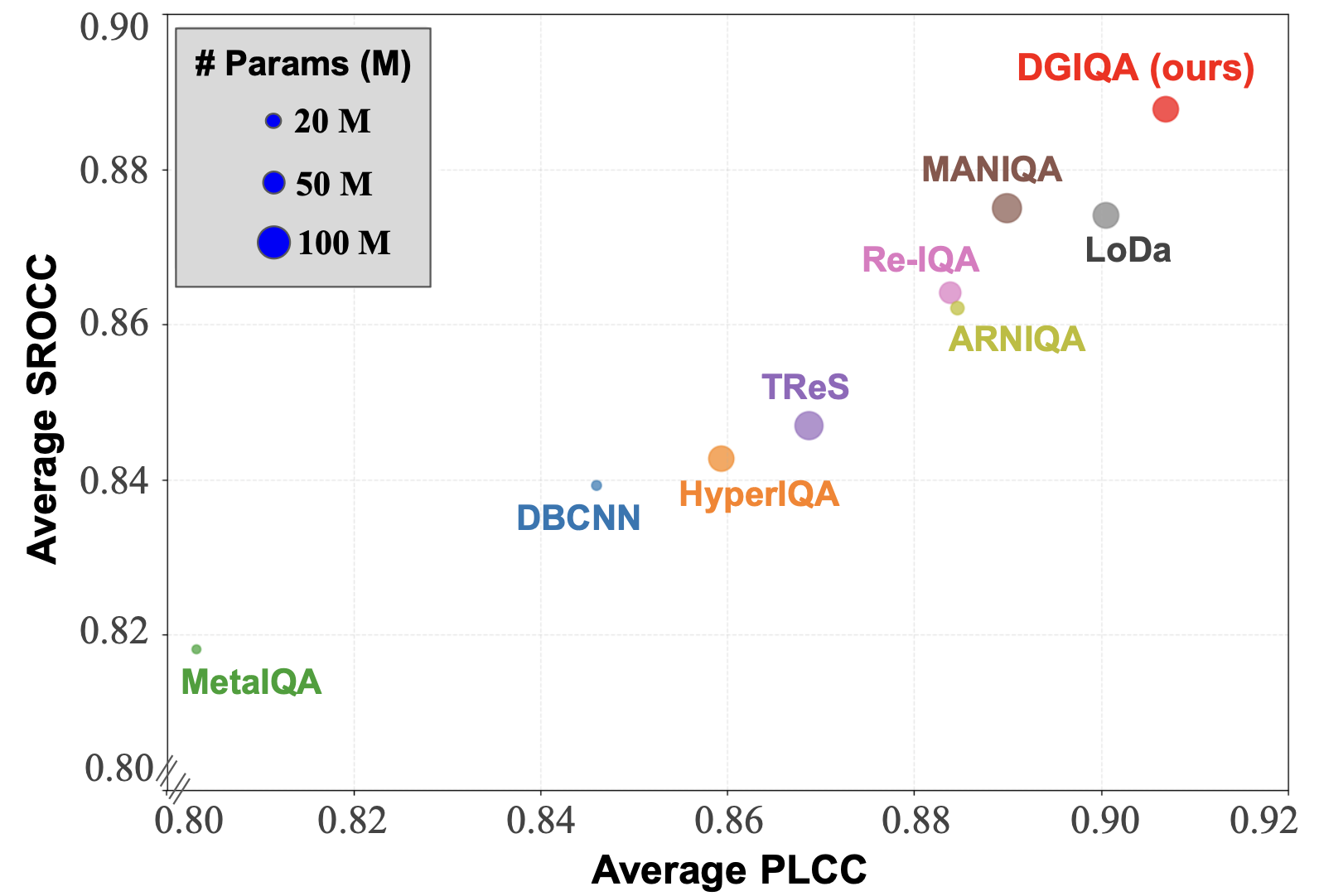}%
    \vspace{-1mm}
    \caption{DGIQA demonstrates improved NR-IQA performance over SOTA models for SROCC (Spearman’s rank order correlation coefficient) and PLCC (Pearson’s linear correlation coefficient) scores averaged over seven benchmark datasets~\cite{wang2021survey,zhai2020perceptual}: LIVE, CSIQ, TID2013, Kadid10k, LIVE-C, Koniq10k, and LIVE-FB.}%
    \label{fig1}
    %\vspace{-1mm}
\end{figure}

With the recent advances in neural networks and deep visual learning, NR-IQA methods have achieved increasingly accurate predictions that align closely with subjective human opinion scores, as shown in Fig.~\ref{fig1}. But they often fail to generalize to \textit{ unseen distortions or out-of-distribution} datasets. This limits their effectiveness in real-world scenarios, where distortions often differ significantly from training conditions. This generalization gap may arise because traditional IQA models rely solely on RGB cues and often fail to attend to the perceptually relevant regions in an image, regions that humans naturally prioritize when judging image quality, such as proximal objects and structural boundaries.

% \begin{figure}[t]
%     \centering
%     \vspace{-1mm}
%     \includegraphics[width=\linewidth]{imgs/Fig1b.png}%
%     \vspace{-1mm}
%     \caption{Comparison of Gaussian Distributions for MOS Scores on the LOL Dataset\cite{Chen2018Retinex}: Probability density plots illustrate the distribution of Mean Opinion Scores (MOS) for high-quality and low-light images. }
%     \label{fig:Fig1b}
% \end{figure}

% Vision Transformers (ViTs)~\cite{vaswani2017attention,dosovitskiy2020image} address this problem by leveraging self-attention mechanisms to extract global contextual relationships~\cite{han2022survey}. In particular, Swin transformers~\cite{liu2021swin} and learnable \textit{quality embeddings}~\cite{cheon2021perceptual} from other transformers have demonstrated remarkable performance for IQA~\cite{yang2022maniqa,ke2021musiq}. Although transformer models induce less locality bias than CNNs~\cite{you2021transformer}, they lack the inductive biases necessary for capturing fine-grained image distortions and structural degradations~\cite{golestaneh2022no}. This is also a common issue in vision-language models~\cite{wu2024comprehensive,zhang2023blind,wang2023exploring}, as they rely on attention-based feature distillation as well. Another major challenge of NR-IQA is the lack of generalization to unseen image distortions~\cite{wang2021survey}, as no \textit{reference} image is available (beyond subjective scores) to guide the learning. 

In human visual perception, depth plays a critical role in guiding attention toward salient, structurally important content, particularly objects that are closer to the observer. Depth cues have been shown to enhance realism and perceived quality of visual experiences in stereoscopic vision~\cite{Hibbard2017-pm}. Beyond perception, depth has been widely used to improve performance in a range of computer vision tasks by providing complementary structural information. In Salient object detection(SOD), additional depth input has shown to boost performance~\cite{rgb-d_salient_chen,10185946,CHEN2019376}. In semantic segmentation, depth contributes to better boundary localization and scene understanding~\cite{10252155,jia2024geminifusionefficientpixelwisemultimodal}.

In this paper, we introduce Depth-guided Cross-Attention and Refinement (Depth-CAR) mechanism to \textit{filter} and \textit{refine} spatial features based on structural guidance of the scene. It performs a spatial attention-based projection to use depth-guided queries to find object saliency and structural information from RGB features for improved IQA predictions. Additionally, to decrease number of parameters without compromising performance, we design the Transformer-CNN Bridge (TCB)-- envisioned to \textit{fuse} the aggregated quality embeddings of a transformer backbone with hierarchical feature representations of CNNs. We implement TCB as an adaptive re-calibration block that learns which feature channels to emphasize, and then use convolution operations to extract their local dependencies.

\begin{figure}[t]
%\vspace{-2mm}
    \centering
    \includegraphics[width=\linewidth]{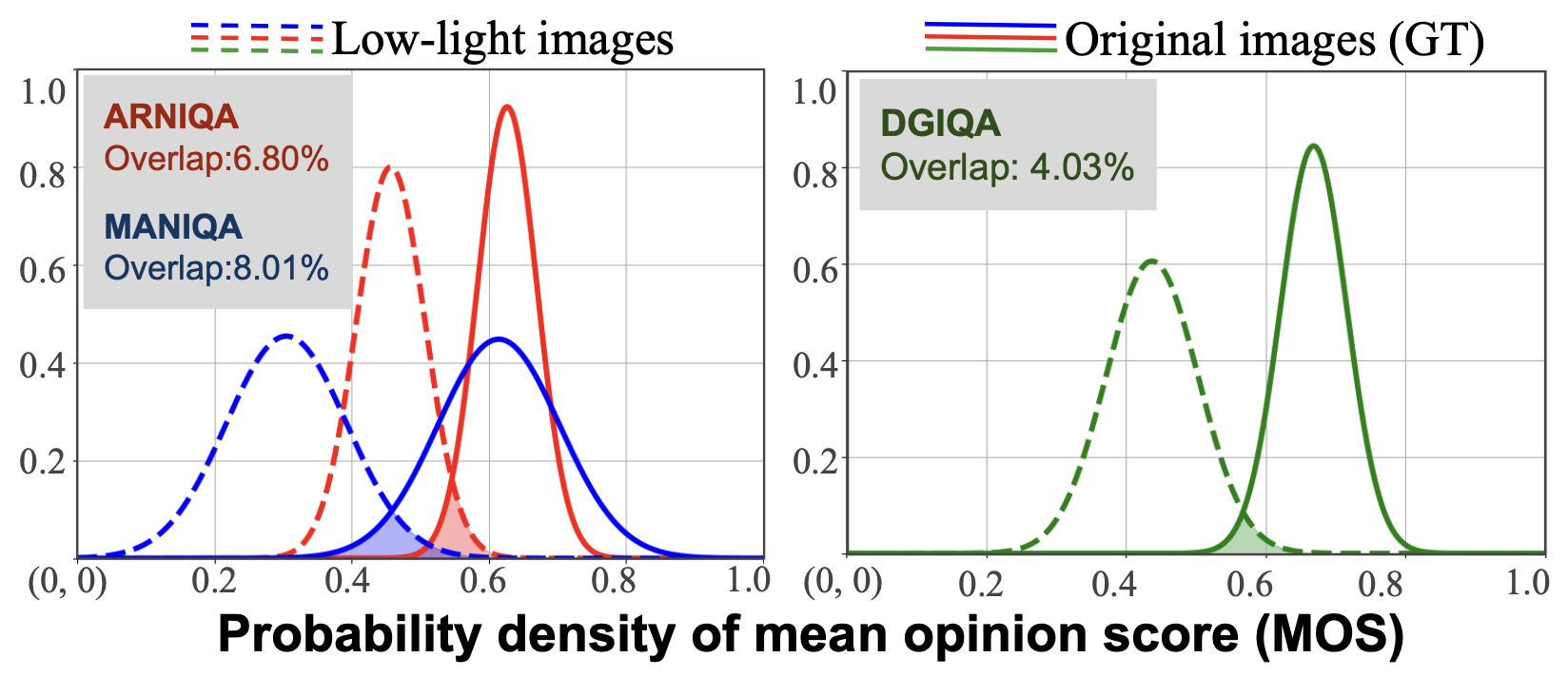}%
    \vspace{-2mm}
    \caption{DGIQA performs $41$\%-$50$\% better separation of IQA score distributions than SOTA models on unseen low-light scenes (LOL dataset~\cite{Chen2018Retinex}); the top transformer model (MANIQA~\cite{yang2022maniqa}) and CNN model (ARNIQA~\cite{agnolucci2024arniqa}) are used for baseline comparison.}%
    \label{fig2}
    %\vspace{-4mm}
\end{figure}

We validate the proposed concepts with comprehensive experiments; our \textbf{DGIQA} model achieves SOTA performance on NR-IQA benchmark datasets of both synthetic (LIVE~\cite{sheikh2006statistical}, CSIQ~\cite{larson2010most}, TID2013~\cite{ponomarenko2015image}, Kadid10k~\cite{lin2019kadid}) and authentic (LIVE-C~\cite{ghadiyaram2015massive}, Koniq10k~\cite{hosu2020koniq} and LIVE-FB~\cite{ying2020patches}) image distortions. Moreover, DGIQA outperforms the SOTA models by considerable margins on cross-dataset evaluations and unseen natural scenes. Our ablation experiments reveal that these improvements are attributed to the multimodal feature learning by TCB blocks and depth-guided feature distillation by Depth-CAR. We further conduct qualitative and quantitative analyses of the generalization performance of DGIQA on low-light imaging and enhancement datasets: LOL~\cite{Chen2018Retinex}, D2G~\cite{KHAN2021115034}, Flare7k~\cite{dai2022flare7k}, IHAZE~\cite{I-HAZE_2018}, and GoPro~\cite{nah2017deep}. These datasets include unseen images with low-light effects, hazy conditions, and lens flares -- which DGIQA can assess and separate over $41\%$ better than SOTA models on all four datasets; see Fig.~\ref{fig2}.

%\JI{write a line about this, cite the datasets and mention the results}
%As shown in Fig.~\ref{fig1}, our DGIQA can assess 41\% - 50\% better in separation on Low-light dataset\cite{Chen2018Retinex} than ARNIQA\cite{agnolucci2024arniqa} and MANIQA.\JI{or the right numbers}

\vspace{1mm}
\noindent
The main contributions of this paper are summarized below:
\begin{enumerate}[label={\arabic*)},nolistsep,leftmargin=*]
\item We present a novel NR-IQA model named \textbf{DGIQA} which incorporrates scene depth to improve NR-IQA learning of structure-level distortions related to focus, blurriness, and other geometric artifacts. We propose a \textit{Depth-CAR} mechanism that enables this by learning a projection function to dynamically refine feature representations based on depth cues, which improves both robustness and efficiency.
\item  We designed a \textit{Transformer-CNN Bridge (TCB)} block that effectively bridges global information of patch embeddings of Transformers with local hierarchical features extracted by CNNs for improved IQA learning.  
\item Comprehensive experiments on benchmark datasets, cross-dataset evaluation, and ablation studies demonstrate the effectiveness of our proposed model and learning pipeline. Specifically, DGIQA achieves SOTA performance on benchmark datasets and outperforms existing models in assessing unseen image distortions. 
\end{enumerate}
We also present a new evaluation criteria for IQA generalization performance, by `\textit{density separation}' of predicted scores on high-quality versus distorted images. As Fig.~\ref{fig2} shows, we demonstrate that unseen data from other imaging domains can be used to validate NR-IQA generalization.

\begin{figure*}[htbp]
    \centering
    %\vspace{-1mm}
    \includegraphics [width=0.96\linewidth]{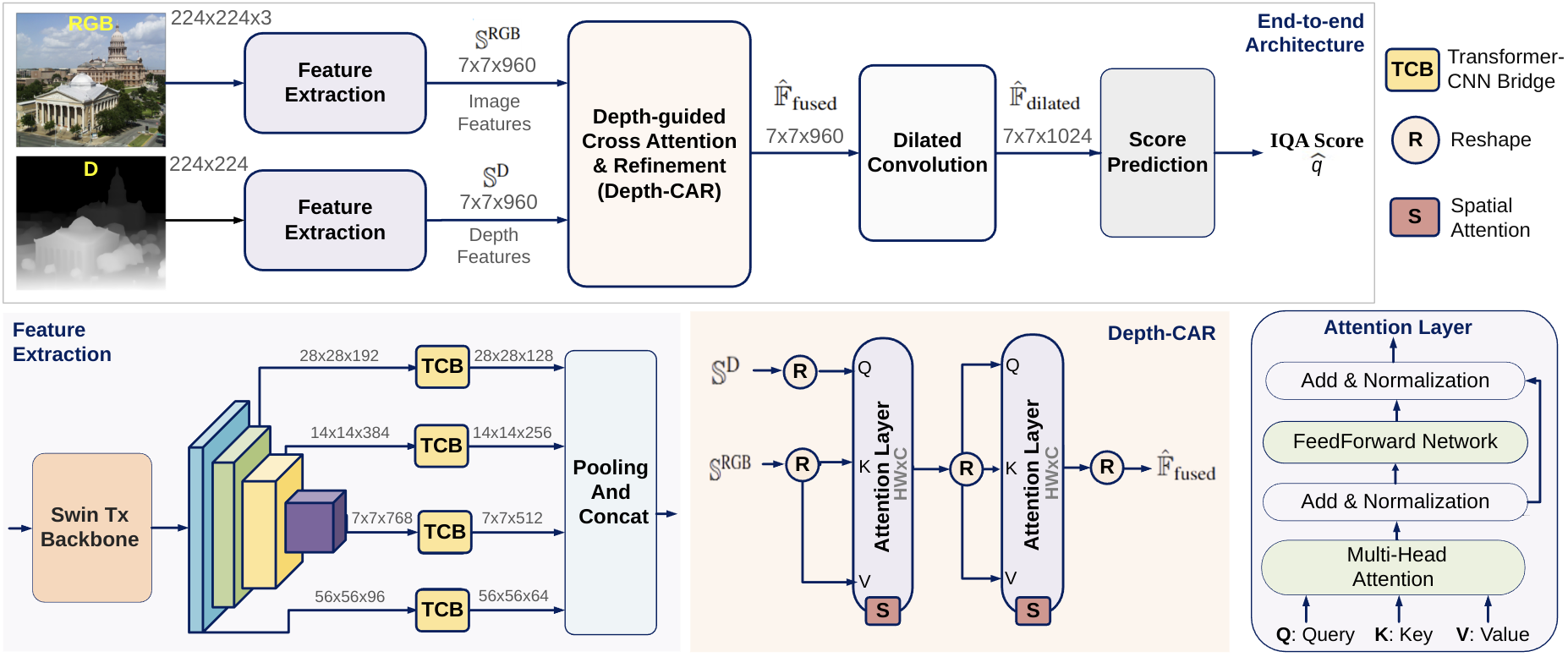}%
    \vspace{-3mm}
    \caption{The model architecture of \textbf{DGIQA} is shown; major learning components are: two pre-trained Swin Transformer backbones for hierarchical feature extraction from input image (RGB) and depth (D); Transformer-CNN Bridge (\textbf{TCB}) blocks for dominant feature distillation; depth-guided cross-attention (\textbf{Depth-CAR}) for RGB-D feature fusion; and a dilated convolution stack -- for NR-IQA learning. 
    }%
    \vspace{-2mm}
\label{fig:model}
\end{figure*}

\section{NR-IQA Literature and Related Work}
\vspace{-1mm}
Classical NR-IQA methods rely on handcrafted features and statistical models for IQA without reference images, with subjective IQA databases as a reliable proxy for objective IQA scores.
Early approaches use the Natural Scene Statistics (NSS)~\cite{gao2013universal,ghadiyaram2017perceptual,mittal2012no,moorthy2011blind, ye2012no,zhang2015feature} to exploit statistical regularities in natural images to detect deviations caused by distortions. BRISQUE (blind image spatial quality evaluator)~\cite{mittal2012no} and BLIINDS (blind image integrity notator using DCT statistics)~\cite{saad2012blind} are prominent metrics that demonstrate robust IQA performance in most scenarios. Besides, HOSA~\cite{xu2016blind} and CORNIA~\cite{ye2012unsupervised} deploy a \textit{code book} list of local features as a reference for guided IQA. These metrics are widely used for their simplicity and efficiency. However, they generally fail to capture complex natural distortions due to limited feature representations and sensitivity to domain shifts. With the advent of deep learning, data-driven approaches~\cite{kang2014convolutional,ma2017end,bosse2017deep} have been dominating the benchmarks for both subjective and objective IQA learning.
%which leads to discrepancies between the measurements and human visual perception.

% \subsection{CNNs and GAN-based Models}
% \vspace{-1mm}

\vspace{1mm}
\noindent
\textbf{CNNs and GAN-based Models}. CNNs are powerful feature extractors that learn hierarchical spatial dependencies from large-scale data, which are crucial to infer image quality~\cite{bianco2018use,lin2018hallucinated,su2020blindly,talebi2018nima,xia2020domain,zhang2018blind,zhu2020metaiqa,jiang2021tongue}. Canonical approaches like CNNIQA~\cite{kang2014convolutional} and WaDiQaM~\cite{bosse2017deep} demonstrated that CNNs outperform statistical approaches in extracting discriminative features for robust IQA. In recent years, HyperIQA~\cite{su2020blindly} reported much-improved performance by learning multi-scale hierarchical features using a context-aware hypernetwork. DBCNN~\cite{zhang2018blind} employed a \textit{seimese} architecture to capture authentic and synthetic distortions, subsequently improving the model's adaptability to various distortions. MetaIQA~\cite{zhu2020metaiqa} proposed a way to learn prior knowledge with two distortion types by meta-learning. Contemporary methods further report better generalization with domain-aware NR-IQA~\cite{xia2020domain} across different distortion levels.

%Inspired by the success of attention models and vision transformers (ViTs) for image recognition tasks, transformer architectures become a new trend in IQA research field. 

% \subsection{Transformer-based Approaches}
% \vspace{-1mm}

\vspace{2mm}
\noindent
\textbf{Transformer-based Approaches}. ViTs efficiently capture long-range dependencies across different parts of an image. IQT~\cite{you2021transformer} demonstrated that transformers induce less locality bias than CNNs and offer better IQA predictions across large databases. MUSIQ~\cite{ke2021musiq} can retain the original image shape by looking up image patches in a 2-D hash map. TReS~\cite{golestaneh2022no} calculates the consistency loss and relative ranking between images and uses a transformer encoder to enhance the final predictions. MANIQA~\cite{yang2022maniqa} achieves SOTA performance by integrating a ViT with a multi-dimensional attention mechanism, which helps capture global contextual information for accurately assessing perceptual image distortions. However, ViTs often lack the inductive biases necessary for capturing local texture details for detecting fine-grained distortions. Vision-language fusion frameworks~\cite{wu2024comprehensive,zhang2023blind,wang2023exploring,chen2025promptiqa} are recently being explored to address this issue, although it remains an open problem.

% \subsection{Knowledge Gaps in the Literature}
% \vspace{-1mm}

\vspace{1mm}

\vspace{-1mm}
\section{Proposed method}
\vspace{-1mm}
\subsection{End-to-end IQA Learning}
\begin{figure}[h]
    \centering
    \vspace{-3mm}
    \includegraphics[width=\linewidth]{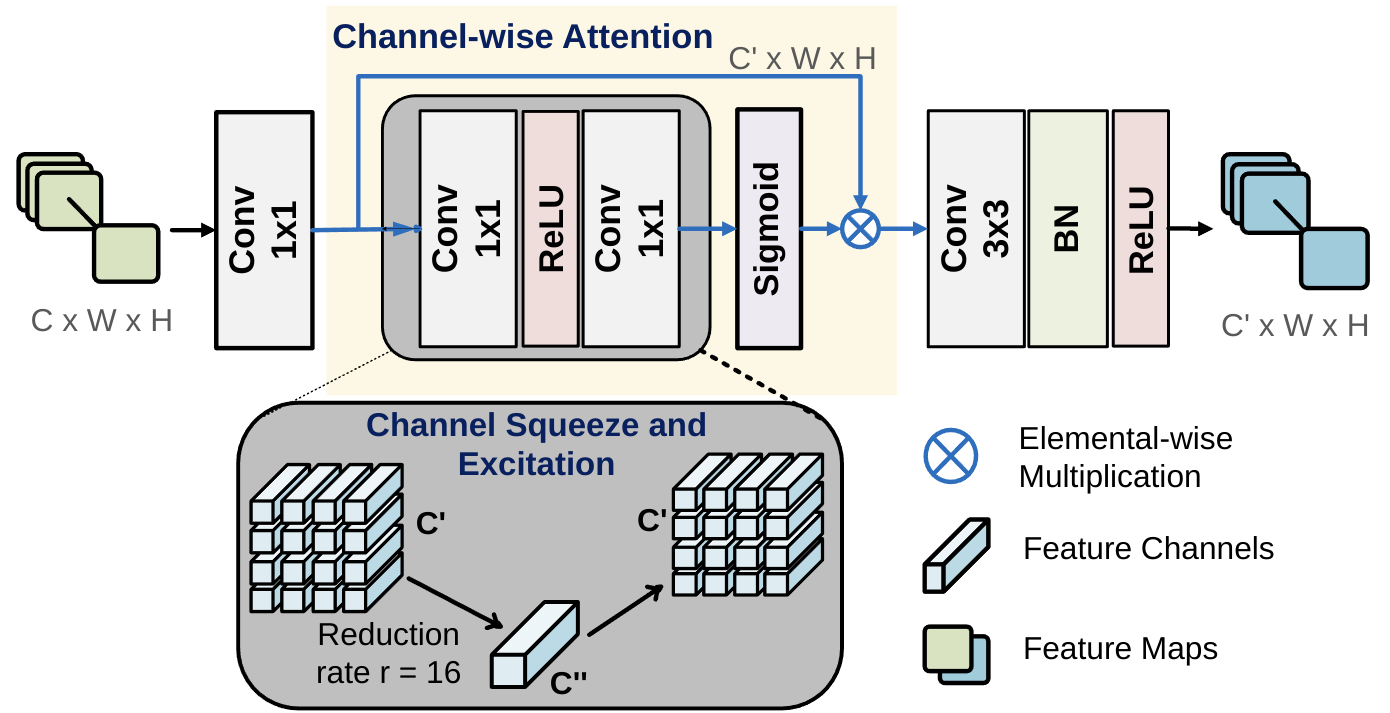}% 
    % Adjust the width to half the text width
    \vspace{-2.5mm}
    \caption{The proposed TCB architecture is shown; it performs a CNN-based projection (resolving local hierarchical dependencies) from the Swin Tx features (capturing global patch-level contexts).}
    %\caption{The proposed TCB (Transformer-CNN Bridge) architecture is shown; it performs a CNN-based projection (resolving local contextual information) from the transformer-extracted features (extracting patch-level dependency).}
    \label{fig:TCB}
    \vspace{-1mm}
\end{figure}

\vspace{-1mm}
The proposed {DGIQA model} is designed to capture both global contextual features and local geometric cues for comprehensive IQA. The end-to-end learning pipeline is presented in Fig.~\ref{fig:model}; given an Image $\mathbf I^{\text{RGB}} \in \mathbb{R}^{H \times W \times 3}$ and corresponding depth map $\mathbf I^{\text{D}} \in \mathbb{R}^{H \times W} $, the objective is to learn its IQA prediction score. We use two swin transformers in parallel to extract hierarchical features for each modality; let the extracted features be denoted as $\mathbb{F_{\text{i}}^{\text{RGB}}}$ and  $ \mathbb{F_{\text{i}}^{\text{D}}}$, respectively where $i$=$1,2,3,4$ indicating the $4$ stages of the backbone feature maps. These feature maps are propagated through four {Transformer-CNN Bridge} ({TCB}) blocks to further extract the local dependencies for distilling the most important features for IQA learning; see Fig.~\ref{fig:TCB}. 
%The refined feature maps can be denoted as $\mathbb{S_{\text{i}}^{\text{RGB}}}$ and  $ \mathbb{S_{\text{i}}^{\text{D}}}$. 

The TCB-distilled features are spatially downsampled, aligned to match dimensions, and concatenated to form the combined feature maps $\mathbb{S^{\text{RGB}}}$ and $\mathbb{S^{\text{D}}}$. We then use a multi-head cross-attention block for depth-guided learning of the object saliency and structural information about the input scene. Specifically, we use the $\mathbb{S^{\text{D}}}$ features as \textit{queries} to extract geometric cues from $\mathbb{S^{\text{RGB}}}$ as \textit{keys}; this mechanism is followed by self-attention for further refinement.

%\textit{channel} and \textit{spatial} attention modules that use the depth features to guide image quality-specific information from the RGB features.  emphasizes geometric cues and helps to identify salient objects relative to depth, producing the fused feature map denoted as $\mathbb{F_{\text{fused}}}$. We use two channel attention blocks to further emphasize the most important \textit{channels} and a \textit{spatial} attention block to condense down the refined feature map $\mathbb{\hat{F}_{\text{fused}}}$.

The refined feature map ($\mathbb{\hat{F}_{\text{fused}}}$) is then passed through a dilated convolution stack~\cite{yu2015multi} that has two {\tt Conv} layers with dilation rates of $2$ and $4$. This helps to expand the receptive field aims to extract multi-scale contextual features from $\mathbb{\hat{F}_{\text{fused}}}$. The last stage applies a global average pooling that compress the final feature maps and propagates it to a fully connected layer for predicting final IQA scores $\hat{q}$.

\subsection{Transformer-CNN Bridge (TCB)}
\vspace{-1mm}
Transformer-based backbones employ attention mechanisms that can capture global dependencies and contextual relevance between distant image patches. Besides, CNNs capture spatial hierarchies and local patterns with progressive convolution layers. Our design intuition for TCB is to take advantage of the transformer features to learn which channels to emphasize, and then use convolution operations to get their local hierarchical dependencies. As shown in Fig.~\ref{fig:TCB}, we envision an adaptive re-calibration of channel-wise features while retaining local spatial information through convolutions. To achieve this, we employ the \textit{squeeze-and-excitation} mechanism~\cite{hu2018squeeze}, where at first, a $1\times1$ {\tt Conv} layer projects the backbone (Swin Tx) extracted features $\mathbb{F}_{i}^{\text{RGB}}$ and $\mathbb{F}_{i}^{\text{D}}$ from channel size $C$=($96$,$192$,$384$,$768$) into $C'$=($64$,$128$,$256$,$512$), respectively; this helps reduce the computational overhead and required training time in the channel attention stage. 

The second $1\times1$ {\tt Conv}  refers to the \textit{squeeze} operation, which projects $C'$ to $C''$=($4$,$8$,$16$,$32$) channels by a reduction rate $r$=$16$. This compresses the information across channels to create a more compact representation; {\tt ReLU} activation is applied to introduce non-linearity. A third $1\times1$ {\tt Conv} is applied for \textit{excitation}, which expands the channel dimension from $C''$ back to $C'$ by learning the respective attention weights. After that, it is passed through a {\tt Sigmoid} (${\sigma}$) to ensure that the attention weights are in [$0$, $1$]. These weights are element-wise multiplied with the original feature map to emphasize the most important channels. While \textit{squeeze-excitation} focuses on channel-wise importance, a subsequent $3\times3$ convolution extracts local spatial dependencies to capture finer details. The filtered features are then passed through batch normalization ({\tt BN}) and {\tt ReLU}, completing the TCB-based feature project process as:
\begin{align}
    \mathbb{A} &= \sigma\big(\text{\tt Conv}_{1 \times 1}(\text{\tt ReLU}(\text{\tt Conv}_{1 \times 1}(\mathbb{F}_{\text{in}})))\big), \\
    \mathbb{F}_{\text{out}} &= \text{\tt ReLU}\big(\text{\tt BN}(\text{\tt Conv}_{3 \times 3}(\mathbb{F}_{\text{in}} \otimes \mathbb{A}))\big).
\label{tcb}
\end{align}
Here, $\mathbb{A}$ is the projected attention map and $\mathbb{F}_{\text{in}}$ ($\mathbb{F}_{\text{out}}$) are the input (output) feature maps, respectively.

\subsection{Depth-guided Cross-Attention \& Refinement}
\vspace{-1mm}
The Depth-CAR block exploits depth features to learn object saliency and structural information from the RGB features for improved IQA predictions. We employ an attention mechanism with queries $Q=\mathbb{S^{\text{D}}}\,\mathbf W_Q$ selected from the depth features, while the keys $K=\mathbb{S^{\text{RGB}}}\,\mathbf W_K$ and values $V=\mathbb{S^{\text{RGB}}}\,\mathbf W_V$ \,that come from the RGB features. Here, $\mathbf W_K$, $\mathbf W_Q$, and $\mathbf W_V$ are learned projection matrices that guide the attention over RGB features to emphasize structural details and object saliency in the input scene.

With this setup, the attention mechanism is computed using scaled dot-product attention given as follows:
\begin{equation} 
\text{\tt Attention}(Q, K, V) = \text{\tt Softmax}\left(\frac{Q K^\top}{\sqrt{d_k}} \right) V 
\end{equation}
where $d_k$ is the embedding dimension. By depth-based queries, our model can focus on closer objects and salient regions in the image, emulating human visual perception that naturally prioritizes nearby prominent object features.

Our depth-guided CAR approach is inspired by recent advancements in the cross-modal attention mechanism for RGB-D feature fusion in semantic segmentation and image registration literature~\cite{10252155,SONG2022102612}. We extend the core idea by using spatial attention to emphasize important $\mathbb{S^{\text{RGB}}}$ features {queried} by $\mathbb{S^{\text{D}}}$ features. We further refine the fused features by self-attention blocks as outlined in the original transformer architecture~\cite{vaswani2017attention}. We use another spatial attention layer for this refinement to capture long-range dependencies within the cross-attended feature maps.

%the fused feature map. The design choice to use more channel attention blocks stems from the fact that channels often contains more information rich features than the individual spatial locations, thus making channel refinement more effective

%Self-attention allows the model to capture long-range dependencies and interactions within the fused feature map. Here we employ two channel attention and a spatial attention to refine the fused feature map. The design choice to use more channel attention blocks stems from the fact that channels often contains more information rich features than the individual spatial locations, thus making channel refinement more effective

%that employs a cross channel attention module to fuse RGB and depth features for enhanced indoor scene semantic segmentation. Similarly, Song et al \cite{SONG2022102612} uses a cross attention layer to fuse multi-modal image inputs for medical image registration.

%The first convolution has a dilation rate of 2, and the second convolution has a dilation rate of 4, allowing the block to capture long-range dependencies and global context from the feature maps while maintaining the spatial resolution

\subsection{Dilation Convolution Block}
\vspace{-1mm}
The dilation convolution block in our model is designed to increase the receptive field of the network without losing resolution. It leverages dilated (atrous) convolutions~\cite{yu2015multi}, which introduce spaces between the {\tt Conv} filter elements, effectively allowing it to cover a wider area without increasing parameters or reducing the spatial dimensions of the feature maps. We employ two consecutive dilated convolutions with dilation rates $2$ and $4$, each followed by batch normalization ({\tt BN}) and {\tt ReLU} operations, as follows: 
\begin{equation}
    \mathbb{\Hat{F}}_{dilated} = \text{\tt ReLU}\big(\text{\tt BN}(\text{\tt Conv}_{3 \times 3, \, dltn=2,4}(\mathbb{\hat{F}}_{fused}))\big).
\end{equation}

\subsection{Score Prediction}
\vspace{-1mm}
Finally, IQA prediction is performed by projecting the $\mathbb{\Hat{F}}_{dilated}$ feature maps into a scalar score $\hat{q}$. After global average pooling condenses the spatial dimensions, the pooled feature vector $\mathbb{F}_{\text{pooled}}$ is passed through a fully connected (FC) layer to generate the predicted score $\hat{q} \in {0,1}$ through a sigmoid activation; specifically, 
\begin{equation}
\hat{q} = \sigma \big(\mathbf W_{\text{fc}} \cdot \mathbb F_{\text{pooled}} + \mathbf b_{\text{fc}} \big)
\end{equation} 
where $ \mathbf{W}_{\text{fc}} $ and $\mathbf b_{\text{fc}} $ are the learnable weights and biases of the FC layer, and $ \sigma $ is the sigmoid function.

\subsection{Loss Functions}
\vspace{-1mm}
We use two loss functions for the supervised training of DGIQA. The primary loss is the Mean-Squared Error (\textbf{MSE}) between the predicted and ground truth scores as: $\mathcal{L}_{\text{MSE}} = \frac{1}{N} \sum_{n=1}^{N} \left({q}_n - \hat{q}_n \right)^2$, where $N$ is the sample batch size. Additionally, a \textbf{Consistency Loss} (CL) inspired by TRES~\cite{golestaneh2022no} is introduced to ensure generalized learning consistency; it measures the MSE between the predicted score of the original image and its horizontally flipped version $\hat{q}_n^{\text{FLIP}}$. Specifically, $\mathcal{L}_{\text{CL}} = \frac{1}{N} \sum_{n=1}^{N} \left( \hat{q}_n - \hat{q}_n^{\text{FLIP}} \right)^2$. With these two loss terms, our objective function is given by
\begin{equation}\label{combined loss}
\mathcal{L}_{\text{Total}} = \mathcal{L}_{\text{MSE}} + \lambda \, \mathcal{L}_{\text{CL}}
\end{equation} 
Here, $\lambda$ is an empirically tuned hyper-parameter that balances the contribution of the consistency loss. %From our experiments, setting $\lambda = 0.5 $, provided us the best results. 

\vspace{-1mm}
\section{Benchmark Evaluation}
\vspace{-1mm}
\subsection{Dataset and Evaluation Criteria}
\vspace{-1mm}
For comprehensive performance evaluation of DGIQA, we perform the experiments on seven benchmark datasets: LIVE\cite{sheikh2006statistical}, CSIQ\cite{larson2010most}, TID2013\cite{ponomarenko2015image}, Kadid10k\cite{lin2019kadid}, LIVE Challenge\cite{ghadiyaram2015massive}, Koniq10k\cite{hosu2020koniq}, and LIVE-FB\cite{ying2020patches}. Each dataset consists of images with varying distortions, and their corresponding ground truth scores, which are derived from subjective estimates by human assessors and are categorized into Mean Opinion Scores (MOS) or Differential Mean Opinion Scores (DMOS). We regularize the scores between $0$ (totally distorted) and $1$ (original image). Table~\ref{DataSummary} shows a summary information on all the datasets.
%For CSIQ, we utilize $866$ reference and distorted images covering six different types of distortions. KADID-10k includes $81$ reference images and $10,125$ distorted images with $25$ types of distortions. KonIQ-10k provides $10,073$ images with authentic distortions and associated MOS scores. LIVE contains $982$ images spanning five distortion types, while TID2013 offers 25 reference images and 3,000 distorted images covering 24 distortion types. 

\begin{table}[ht]
\centering
\caption{Information on benchmark datasets used in the training and evaluation of DGIQA.}
\vspace{-3mm}
\resizebox{\linewidth}{!}{  % Resizes the table to fit within the width of the page
\begin{tabular}{llrrl}
\toprule
\textbf{Dataset}       & \textbf{Types}    & \textbf{\# Images} & \textbf{\# Distortions} & Scores \\
\midrule
LIVE                   & Synthetic         & 982               & 5    & DMOS     \\
CSIQ                   & Synthetic         & 866               & 6   &   DMOS     \\
TID2013                & Synthetic         & 3,000             & 24  &  MOS     \\
Kadid10k               & Synthetic         & 10,125            & 25   &  DMOS    \\
Live Challenge         & Authentic         & 1,162             & -    &   MOS   \\
Koniq10k               & Authentic         & 10,073            & -    &    MOS  \\
LIVE-FB                & Authentic         & 39,810            & -    &    MOS  \\
\bottomrule
\end{tabular}
}
\label{DataSummary}
\vspace{-2mm}
\end{table}

% \vspace{1mm}
\noindent
\textbf{Evaluation Metrics}. We use two standard metrics~\cite{wang2004image,zhai2020perceptual}: Spearman’s rank order correlation coefficient (SROCC) and Pearson’s linear correlation coefficient (PLCC) to evaluate the correlation between the predicted scores and the true scores of the distorted pictures. PLCC is defined as:
\begin{equation*}
PLCC=\frac{\sum_{i=1}^N(\hat{q}_i-\Bar{\hat{q}})(q_i-\Bar{q})}{\sqrt{\sum_{i=1}^N(\hat{q}_i-\Bar{\hat{q}})^2}\sqrt{\sum_{i=1}^N(q_i-\Bar{q})^2}}
\end{equation*}
where $N$ is the number of distorted images, $\hat{q}_i$ is the predicted scores, $q_i$ is the true scores, and $\Bar{\hat{q}}$ and $\Bar{q}$ are the corresponding means. 
Besides, SROCC is defined as:
\begin{equation*}
SROCC=1-\frac{6\sum^N_{i=1}d^2_i}{N(N^2-1)}
\end{equation*}
where $d_i$ is the pairwise differences between ${q}_i$ and $\hat{q}_i$. 

%\begin{comment}
\begin{table*}[ht]
\centering
\caption{SROCC and PLCC results comparison of DGIQA against SOTA NR-IQA models on synthetic and authentic datasets are shown. Top two scores are marked in {\color{blue}blue}; the best scores are \textbf{\color{blue}bold}; `-' indicates the model checkpoints and results are not available/reported.}
\vspace{-2mm}
\resizebox{\textwidth}{!}{  % Resizes the table to fit the text width
\begin{tabular}{l||cccccccc|cccccc}
\Xhline{2\arrayrulewidth}
& \multicolumn{8}{c|}{\textbf{Synthetic Datasets}} & \multicolumn{6}{|c}{\textbf{Authentic Datasets}} \\
\cline{2-9} \cline{10-15}
& \multicolumn{2}{c}{LIVE} & \multicolumn{2}{c}{CSIQ} & \multicolumn{2}{c}{TID2013} & \multicolumn{2}{c|}{Kadid10k} 
& \multicolumn{2}{|c}{Live-Challenge} & \multicolumn{2}{c}{Koniq10k} & \multicolumn{2}{c}{LIVE-FB} \\
\cline{2-3} \cline{4-5} \cline{6-7} \cline{8-9} \cline{10-11}
\cline{12-13} \cline{14-15}
\textbf{Model} & \textbf{SROCC} & \textbf{PLCC} & \textbf{SROCC} & \textbf{PLCC} & \textbf{SROCC} & \textbf{PLCC} & \textbf{SROCC} & \textbf{PLCC} & \textbf{SROCC} & \textbf{PLCC} & \textbf{SROCC} & \textbf{PLCC} & \textbf{SROCC} & \textbf{PLCC} \\
\Xhline{2\arrayrulewidth}
DIIVINE~\cite{saad2012blind}  & 0.892 & 0.908 & 0.804 & 0.776 & 0.643 & 0.567 & 0.413 & 0.435 & 0.588 & 0.591 & 0.546 & 0.558 & 0.092 & 0.187 \\
BRISQUE~\cite{mittal2012no}  & 0.929 & 0.944 & 0.812 & 0.748 & 0.626 & 0.571 & 0.528 & 0.567 & 0.629 & 0.629 & 0.681 & 0.685 & 0.303 & 0.341 \\
% WaDIQaM~\cite{bosse2017deep}  & 0.960 & 0.955 & 0.852 & 0.844 & 0.835 & 0.855 & 0.739 & 0.752 & 0.682 & 0.671 & 0.804 & 0.807 & 0.455 & 0.467 \\
% DBCNN~\cite{zhang2018blind}  & 0.968 & 0.971 & 0.946 & 0.959 & 0.816 & 0.865 & 0.856 & 0.851 & 0.869 & 0.869 & 0.875 & 0.884 & 0.545 & 0.551 \\
HyperIQA~\cite{su2020blindly}  & 0.962 & 0.966 & 0.923 & 0.942 & 0.840 & 0.858 & 0.852 & 0.845 & 0.872 & 0.885 & 0.906 & 0.917 & 0.544 & 0.602 \\
% MetalIQA~\cite{zhu2020edge}& 0.960 & 0.959 & 0.856 & 0.868 & 0.762 & 0.775 & 0.887 & 0.856 & 0.835 & 0.802 & 0.887 & 0.856 & 0.540 & 0.507 \\
MUSIQ~\cite{ke2021musiq} & 0.940 & 0.911 & - & - & 0.773 & 0.815 & 0.875 & 0.872 & 0.702 & 0.746 & 0.916 & 0.928 & 0.566 & 0.661 \\
TReS~\cite{golestaneh2022no}  & 0.969 & 0.968 & 0.922 & 0.942 & 0.863 & 0.883 & 0.859 & 0.858 & 0.846 & 0.877 & 0.915 & 0.928 & 0.554 & 0.625 \\
MANIQA~\cite{yang2022maniqa}  & \textbf{\color{blue}0.982} & \textbf{\color{blue}0.983} & \color{blue}0.968 & 0.961 & \textbf{\color{blue}0.937 }& \textbf{\color{blue}0.943} & \color{blue}{0.944} & \color{blue}0.943 & 0.871 & 0.887 & 0.880 & 0.915 & 0.543 & 0.597 \\
%Su et al.\cite{su2023distortion}  & 0.973 & 0.974 & 0.935 & 0.952 & 0.815 & 0.859 & 0.866 & 0.874 & - & - & - & - & - & - \\
Re-IQA~\cite{saha2023re}& 0.970 & 0.971 & 0.947 & 0.960 & 0.804 & 0.861 & 0.872 & 0.885 & 0.840 & 0.854 & 0.914 & 0.923 & \color{blue}0.645 & \color{blue}0.733 \\
LoDa~\cite{xu2024boosting}  & 0.975 & 0.979 & 0.958 & 0.965 & 0.869 & 0.901 & 0.931 & 0.936 & 0.876 & 0.899 & \color{blue}0.932 & \textbf{\color{blue}0.944} & 0.578 & 0.679 \\
ARNIQA~\cite{agnolucci2024arniqa}  & 0.966 & 0.972 & 0.962 & \color{blue}0.973 & 0.880 & 0.901 & 0.908 & 0.912 & - & - & - & - & \color{blue}{0.595} & 0.671 \\
LIQE~\cite{zhang2023liqe}  & 0.970 & 0.951 & 0.936 &  0.939 & - & - & 0.931 & 0.930  & \textbf{\color{blue}0.904} & \textbf{\color{blue}0.910} & 0.919  & 0.908 & - & - \\
$\text{CLIPIQA}^{+}$~\cite{wang2023exploring}  & 0.948 & 0.952  & 0.907 & 0.928  & 0.835 & 0.857  & 0.913 & 0.909  & 0.805 & 0.832  &  0.895 & 0.909   &  0.540 & 0.566  \\
TOPIQ~\cite{chen2023topiqtopdownapproachsemantics} & - & - & - & - & - & -  & -  & - & 0.870  & 0.884 & 0.926 & 0.939 & \textbf{\color{blue}0.652}  & \textbf{\color{blue}0.745} \\
%SaTQA\cite{shi2024transformer} & - & - & \color{blue}0.965 & \color{blue}0.972 & \textbf{\color{blue}0.938} & \textbf{\color{blue}0.948} & \textbf{\color{blue}0.946} & \textbf{\color{blue}0.949} & \color{blue}0.877 & \color{blue}0.903 & \color{blue}0.930 & \color{blue}0.941 & \color{blue}0.582 & \color{blue}0.676 \\
\textbf{DGIQA~(ours)} & \textbf{\color{blue}0.982} & \color{blue}0.981 & \color{blue}\textbf{0.971} & \textbf{\color{blue}0.976} & \color{blue}0.934 & \color{blue}0.940 & \color{blue}0.943 & \textbf{\color{blue}0.945} & \color{blue}0.891 & \textbf{\color{blue}0.910} & \textbf{\color{blue}0.934} & \color{blue}0.942 & 0.591 & 0.685 \\
\Xhline{2\arrayrulewidth}
\end{tabular}
}
\label{results}
\vspace{-2mm}
\end{table*}

\vspace{-1mm}
\subsection{Implementation details}\label{imp details}
\vspace{-1mm}
We implement DGIQA using PyTorch libraries, with a NVIDIA A100 GPU for training and evaluation. Following standard practices~\cite{yang2022maniqa,zhu2020metaiqa}, we address varying input sizes by taking random $224\times224$ crops from the image and depth map instead of resizing during training. The DepthAnything model~\cite{yang2024depth} is used for generating depth pairs when ground truth is not available. Data augmentation includes horizontal and vertical flips applied uniformly (with a probability of $0.5$). We use the albumentation library~\cite{info11020125} to ensure RGB-D alignment during data augmentation. 
%For tests, we evaluate IQA models on $10$ random $224\times224$ crops from each image and depth map; their mean score is reported as the final prediction. 

The two Swin Transformer (Swin-T) backbones~\cite{liu2021swin} of DGIQA are pre-trained on ImageNet-21k dataset to (learn to) extract multi-scale feature maps from RGB and depth inputs. We use AdamW optimizer with a weight decay of $1e^{-2}$ and a learning rate of $1e^{-4}$ ($1e^{-5}$) for synthetic (authentic) datasets. The learning rate schedule follows a Cosine Annealing strategy~\cite{yang2024align} where $T_{max}$ is set to half the number of total epochs at which point the learning rate reaches its minimum value of $1e^{-7}$. The maximum training duration is set to $200$ epochs. The batch size is set to $16$ to balance the memory cost and training stability. Following the usual IQA training practice, the dataset is split $80$:$20$, repeated $10$ times with different seeds; the mean SROCC and PLCC scores are then reported for evaluation.

\subsection{Performance Comparison With SOTA}
\noindent
\textbf{Dataset Evaluation}. The benchmark datasets (Table~\ref{DataSummary}) include a diverse range of images with various types and levels of distortions, enabling a comprehensive objective evaluation of how closely a model’s predictions align with human perception. Table~\ref{results} presents the performance of our DGIQA model in terms of SROCC and PLCC, compared to SOTA models across synthetic and authentic datasets. Our model consistently ranks within the \textbf{top two} and notably achieves the \textbf{best performance} on LIVE, CSIQ, Kadid10k, LIVE-C, and Koniq10k datasets. Such consistent scores across these diverse datasets validate the robustness and effectiveness of DGIQA learning pipeline. %In particular, the integration of depth information enhances the model's ability to capture visual quality distortions in real-world scenarios, making it a strong candidate for image quality assessment tasks across both synthetic and authentic datasets.

\vspace{1mm}
\noindent
\textbf{Cross-dataset Evaluation}. In many real-world applications, image quality distortions can vary significantly from those seen during training. Cross-dataset evaluation helps examine how well an NR-IQA model performs on entirely different datasets with unique images, distortion types, and quality levels. This evaluation approach helps reveal overfitting tendencies as well. We evaluate the performance of DGIQA through comprehensive cross-dataset evaluations on both synthetic and authentic datasets. Results are reported in Table~\ref{tab:cross_dataset_result} showing that our model \textbf{outperforms the SOTA models} in most dataset pairs. Specifically, DGIQA achieves $0.79$\%-$2.80$\% ($3.86$\%-$6.41$\%) improvements on SROCC (PLCC) scores on authentic datasets. The corresponding improvements on synthetic datasets are $2.64$\%-$4.96$\% ($0.41$\%-$4.78$\%) for SROCC (PLCC) scores, respectively. As Table~\ref{tab:cross_dataset_result} indicates, our DGIQA exhibits a strong generalization across diverse scenes and distortion types, verifying its robustness and learning effectiveness.

\begin{table}[h!]
\caption{Cross-dataset performance of DGIQA is shown in comparison with the closest competitors: MANIQA, LoDa, and other models that report good cross-validation IQA results; `-' indicates the model checkpoints and results are not available/reported. }%
\vspace{-2mm}
\small
\resizebox{\columnwidth}{!}{%
\begin{tabular}{l|cc|cc|cc}
\multicolumn{7}{c}{{\normalsize (\textbf{a}) Evaluation on authentic datasets.}} \\
\Xhline{2\arrayrulewidth}
\multicolumn{1}{r|}{Train On} & \multicolumn{4}{c|}{\textbf{Koniq10K }}  & \multicolumn{2}{c}{\textbf{LIVE-FB}}  \\
\Xhline{2\arrayrulewidth}
\multicolumn{1}{r|}{Test On} & \multicolumn{2}{c|}{\textbf{LIVE-C}} & \multicolumn{2}{c|}{\textbf{LIVE-FB}} & \multicolumn{2}{c}{\textbf{Koniq10K }}  \\
Model $\downarrow$ & SROCC & PLCC & SROCC & PLCC & SROCC & PLCC  \\
\Xhline{2\arrayrulewidth}
TReS~\cite{golestaneh2022no} & 0.786 & \color{blue}{0.803} & \color{blue}{0.453} &  \color{blue}\textbf{0.559} & 0.703 & 0.713 \\
MANIQA~\cite{yang2022maniqa} &0.792 & 0.807  & 0.405 & 0.438 & - & -\\
% Re-IQA$^*$ \cite{saha2023re} &  & &  &\\
ARNIQA~\cite{agnolucci2024arniqa} & 0.723  & 0.758 & 0.410 & 0.486  & 0.738 & \color{blue}0.749  \\ 
LoDa~\cite{xu2024boosting}  & \color{blue}\textbf{0.811} & - & - & - & \color{blue}0.763 & - \\
%DGIQA (w/o D-CAR) & 0.783  & 0.816 &  &  \\
%DGIQA (w/o Dilation) & 0.804 & 0.830  &  &   \\
%DGIQA (w/o TCB) & 0.806 & 0.828 &  &  \\
\textbf{DGIQA} & \color{blue}0.808 & \color{blue}\textbf{0.834} & \color{blue}\textbf{0.462} & \color{blue}{0.515} & \color{blue}\textbf{0.769}&\color{blue}\textbf{0.797}\\
\Xhline{2\arrayrulewidth}
\end{tabular}%
}

\vspace{2mm}
\resizebox{\columnwidth}{!}{%
\begin{tabular}{l|cc|cc|cc}
\multicolumn{7}{c}{{\normalsize (\textbf{b}) Evaluation on synthetic datasets.}} \\
\Xhline{2\arrayrulewidth}
\multicolumn{1}{r|}{Train On} & \multicolumn{4}{c|}{\textbf{ Kadid10k}}  & \multicolumn{2}{c}{\textbf{TID2013}} \\
\Xhline{2\arrayrulewidth}
\multicolumn{1}{r|}{Test On} & \multicolumn{2}{c|}{\textbf{ LIVE}} & \multicolumn{2}{c|}{\textbf{TID2013}} & \multicolumn{2}{c}{\textbf{LIVE}} \\
Model $\downarrow$ & SROCC & PLCC & SROCC & PLCC & SROCC & PLCC  \\
\Xhline{2\arrayrulewidth}
HyperIQA~\cite{su2020blindly}  & \color{blue}{0.908} & - & 0.706 &  - & 0.876 & - \\
TReS~\cite{golestaneh2022no} & 0.894 & 0.872 & 0.701 & 0.726 & 0.857 & 0.842 \\
MANIQA~\cite{yang2022maniqa} &0.865 & 0.820  & 0.700 & \color{blue}0.730 & - & -\\
% Re-IQA$^*$ \cite{saha2023re} & 0.892 & & \color{blue}\textbf{0.777} &\\
ARNIQA~\cite{agnolucci2024arniqa} & 0.898  & \color{blue}{0.900} & \textbf{\color{blue}{0.760} }& 0.715  & \color{blue}{0.887} & \color{blue}{0.880}\\ 
% DGIQA (w/o TCB) &  &  &  &  \\
% DGIQA (w/o D-CAR) &  &  &  &   \\
% DGIQA (w/o Dilation) &  &  &  &  \\
\textbf{DGIQA} & \color{blue}\textbf{0.932} & \color{blue}\textbf{0.909} & \color{blue}0.724 & \color{blue}\textbf{0.733} & \color{blue}\textbf{0.931} & \color{blue}\textbf{0.922} \\
\Xhline{2\arrayrulewidth}
\end{tabular}%
}
\label{tab:cross_dataset_result}
%\vspace{-4mm}
\end{table}

\vspace{-1mm}
\subsection{Ablation Experiments}
\begin{figure}[h]
%\vspace{-1mm}
\centering
\includegraphics[width=\linewidth]{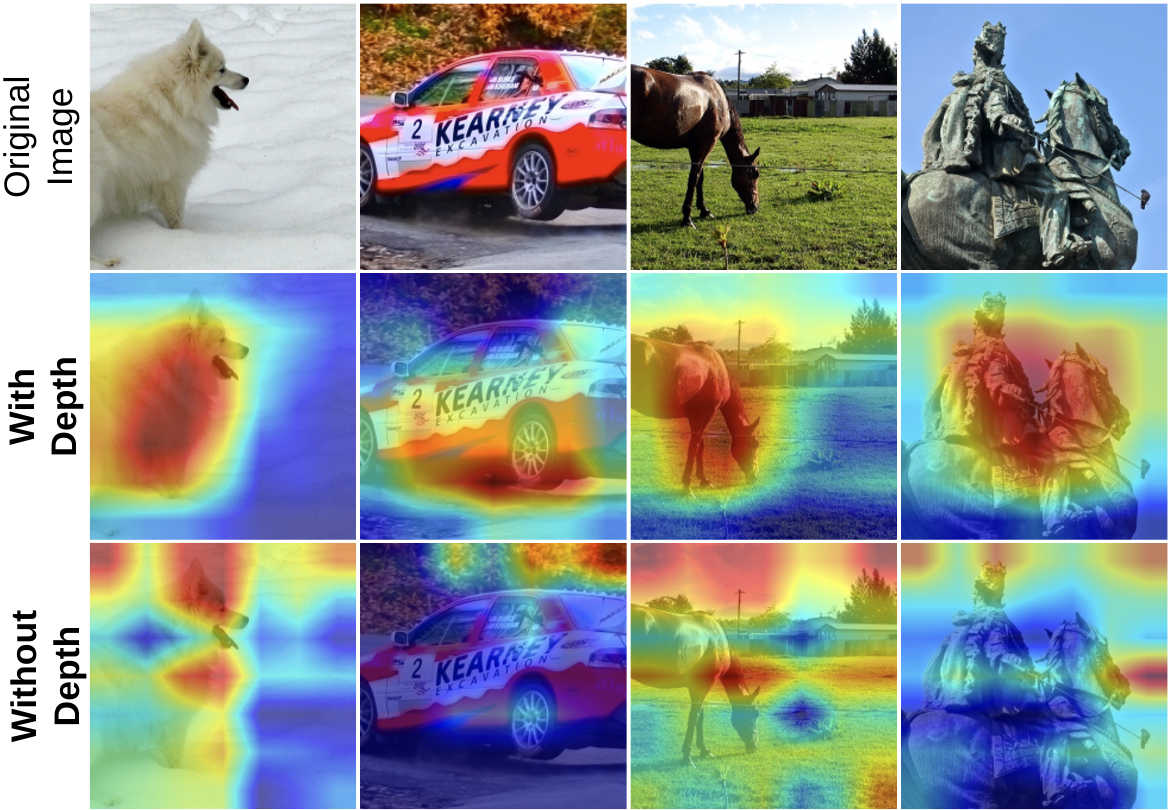}%
\vspace{-2.5mm}
\caption{A few instances of Grad-CAM~\cite{8237336} visualizations of DGIQA model's attention (from the dilated stack) are shown, demonstrating more guided attention on salient objects when depth is used for training, compared to using RGB inputs alone.}
%\vspace{-2mm}
\label{fig:gradcam}
\end{figure}

\vspace{-1mm}
We perform several ablation studies to demonstrate the effectiveness of major computational components of DGIQA. We analyze our model performance with and without TCB blocks, Depth-CAR, and dilated convolution blocks -- in comparison with the full model. Fig.~\ref{fig:gradcam} illustrates the effectiveness of our multimodal learning, Table~\ref{ablation_study} presents the results for the LIVE-C, and Table~\ref{cross_result_ablations} presents the cross-dataset validation on LIVE-C and LIVE-FB datasets with models trained on the Koniq10K dataset alone.

\begin{table}[h!]
\vspace{-1mm}
\centering
\caption{Ablation results on LIVE-C dataset are shown for four configurations of the proposed DGIQA model.}
\vspace{-3mm}
\resizebox{\columnwidth}{!}{%
\begin{tabular}{c|ccc|cc|c}
\Xhline{2\arrayrulewidth}
Cfg & TCB & Depth-CAR & Dilation & SROCC & PLCC & \# Params\\ 
\Xhline{2\arrayrulewidth}
% 1 & \checkmark &  & \checkmark & 0.9280 & 0.9312 & $65$\,M\\ 
% 2 &\checkmark  & \checkmark &  & 0.9292 & 0.9385 & $103$\,M\\ 
% 3 &   & \checkmark & \checkmark  & 0.9300 & 0.9394 & $127$\,M\\ 
% 4 & \checkmark & \checkmark & \checkmark & \textbf{0.9311} & \textbf{0.9400} & $103$\,M\\ 
\#1 & \checkmark &  & \checkmark & 0.877 & 0.902 & $65$\,M\\ 
\#2 &\checkmark  & \checkmark &  & 0.888 & 0.907 & $103$\,M\\ 
\#3 &   & \checkmark & \checkmark  & 0.890 & 0.907 & $127$\,M\\ 
\#4 & \checkmark & \checkmark & \checkmark & \textbf{0.891} & \textbf{0.909} & \textbf{103}\,M\\ 
\Xhline{2\arrayrulewidth}
\end{tabular}%
}
\label{ablation_study}
%\vspace{-4mm}
\end{table}

\vspace{1mm}
\noindent
\textbf{Effects of Depth Fusion}. To demonstrate the effectiveness of incorporating depth information in our learning pipeline, we use Grad-Cam~\cite{8237336} to generate heatmaps that visualize regions where DGIQA focuses for IQA predictions. We show a few comparisons in Fig.~\ref{fig:gradcam}, which clearly illustrates that depth information guides our model to focus on salient objects~\cite{islam2022svam}, leading to more accurate predictions that align closely with how human visual perception works. In contrast, without depth information, the model's attention is more dispersed and less targeted.

Table~\ref{ablation_study} further demonstrates that {Depth-Guided CAR} improves DGIQA performance by $1.6$\% ($0.8$\%) in SROCC (PLCC). It also boosts the cross-dataset performance by $3.2$\% ($2.2$\%) on LIVE-C, and $8.2$\% ($5.3$\%) on LIVE-FB in SROCC (PLCC), respectively; see Table~\ref{cross_result_ablations}.

\begin{table}[h!]
\vspace{-2mm}
\caption{Cross-dataset ablation results of DGIQA trained on Koniq10K alone (for the same 4 configurations of Table~\ref{ablation_study}).} 
\vspace{-1mm}
\small
\resizebox{\columnwidth}{!}{%
\begin{tabular}{l|cc|cc}
\hline
\multicolumn{1}{r|}{\textbf{Test On} $\rightarrow$}  & \multicolumn{2}{c|}{\textbf{LIVE-C}} & \multicolumn{2}{c}{\textbf{LIVE-FB}}  \\
Model Configurations $\downarrow$ & SROCC & PLCC & SROCC & PLCC   \\
\hline
%HyperIQA \cite{su2020blindly}  & 0.785 & 0.746 & 0.437 & 0.465  \\
%TReS \cite{golestaneh2022no} & \underline{0.786} & \underline{0.803} & \textbf{0.453} &  \textbf{0.559} \\
%MANIQA \cite{yang2022maniqa} &0.753 & 0.758  & 0.358 & 0.398  \\
%Re-IQA \cite{saha2023re} &  & &  &\\
%ARNIQA \cite{agnolucci2024arniqa} & 0.723  & 0.758 & 0.410 & 0.486    \\ \hline

\#1.~DGIQA (w/o D-CAR) & 0.783  & 0.816 & 0.427 & 0.489 \\
\#2.~DGIQA (w/o Dilation) & 0.804 & 0.830  & 0.431 & 0.490  \\
\#3.~DGIQA (w/o TCB) & 0.806 & 0.828 & 0.457& 0.509 \\
\#4.~\textbf{DGIQA (Full)} & \textbf{0.808} & \textbf{0.834} & \textbf{0.462} & \textbf{0.515} \\
\bottomrule
\end{tabular}%
}
\label{cross_result_ablations}
\vspace{-4mm}
\end{table}

\vspace{1mm}
\noindent
\textbf{Impacts of TCB Blocks}. As mentioned earlier, TCB blocks are envisioned to exploit the Swin Tx features to capture more local interactions with the subsequent {\tt Conv} operations. Then, it prioritizes the relevant channels for IQA predictions by its squeeze-and-excitation mechanism. We observe that these process downsamples feature maps, thus reducing the learnable parameters without compromising performance. As Table~\ref{ablation_study} and Table~\ref{cross_result_ablations} demonstrate, the addition of TCB blocks increases overall NR-IQA performance, while \textbf{reducing the model parameters by 23.3\%}. Effectively, it saves one-fifth of the training and inference time in addition to boosting net NR-IQA performance.

\vspace{1mm}
\noindent
\textbf{Contributions of Dilated Convolution}. Dilated convolutions introduce gaps between pixels in the convolutional filters, effectively widening the receptive field. This enables the model to capture information from a broader context. As Table~\ref{cross_result_ablations} demonstrates, dilation stacks increase overall performance by $0.5$\% ($0.5$\%) on LIVE-C and $7.2$\% ($5.1$\%) on LIVE-FB dataset in SROCC (PLCC) scores, respectively.

\begin{figure}[h]
%\vspace{-1mm}
    \centering
    \includegraphics[width=0.98\linewidth]{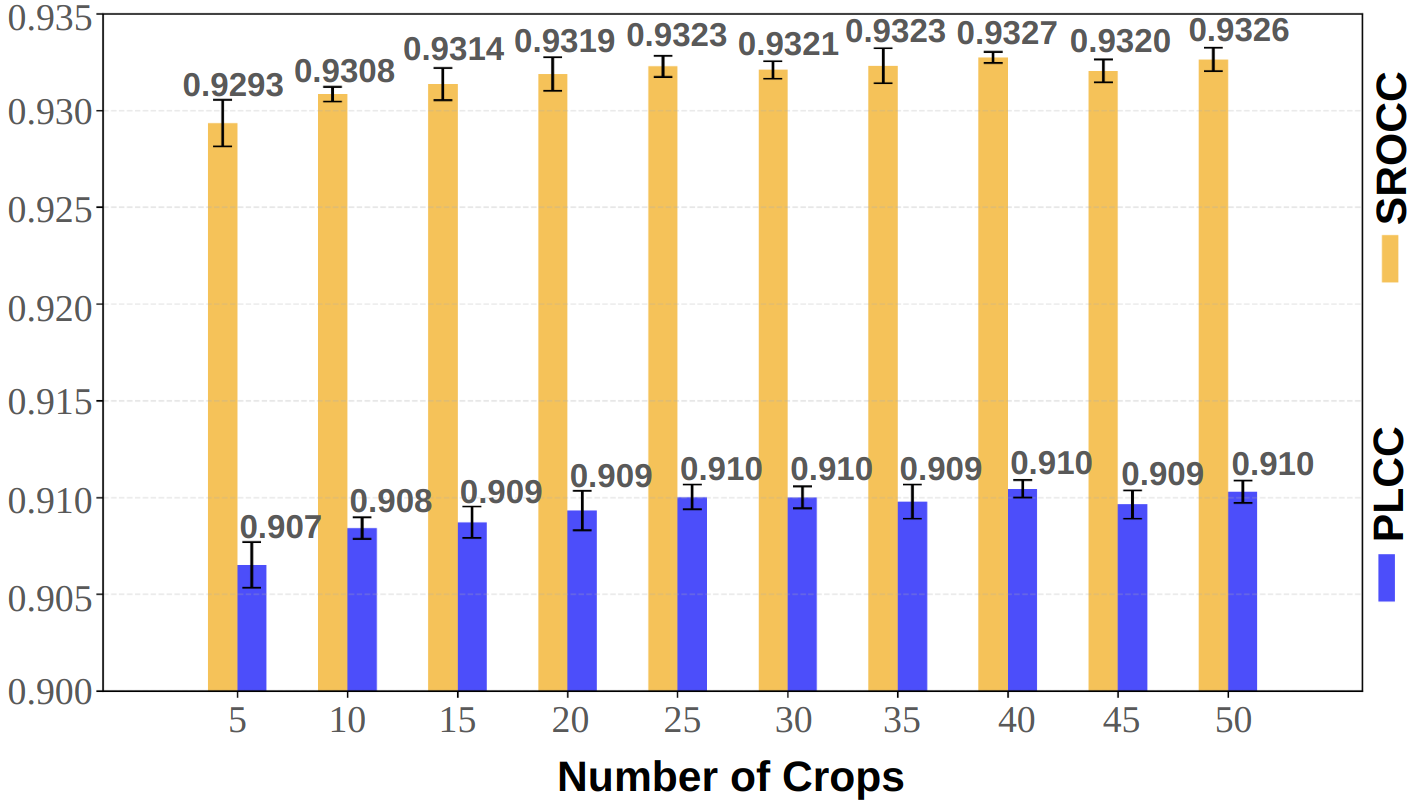}%
    \vspace{-2mm}
    \caption{Cross-dataset ablation results based on different numbers of crops, trained on Kadid10k and tested on LIVE dataset. }
    \label{fig:crops}
    %\vspace{-2mm}
\end{figure}

\vspace{0mm}
\noindent
\textbf{Impacts of Hyper-parameters}. Fig.~\ref{fig:crops} presents the cross-validation performance with a varied number of crops selected for evaluation. We select $25$ crops as a balance between performance and validation time. We also investigate the impacts of the contribution factor of the consistency loss, $\lambda$ (see Eq.~\ref{combined loss}); $0.3$ gives the best results while corner values towards $0$ and $1$ reduce the performance. More detailed results are in Appendix~\ref{sec:Appendix-HyperparameterTuning}. 

%we tried different set of $\lambda$, $0.3$ gives the best results while other values from $0$ to $1$ slightly reduce the performance. 

\vspace{-2mm}
\section{Generalized IQA Learning}
\subsection{Effective Feature Distillation}
\begin{figure}[t]
\centering
\begin{subfigure}{\linewidth}
    \centering
    \includegraphics[width=0.98\linewidth]{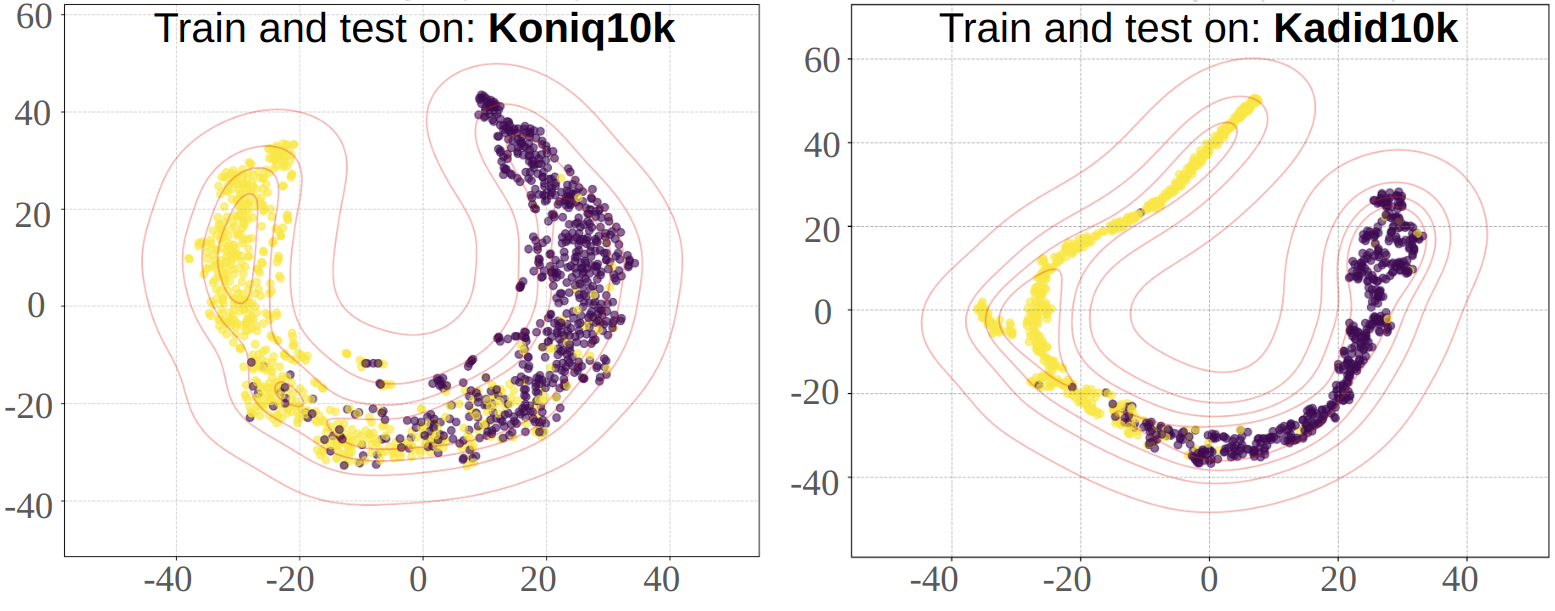}%
    \vspace{-1mm}
    \caption{Benchmark evaluation on Koniq10k and Kadid10k datasets.}
    \label{fig:tsne_top}
\end{subfigure}

\begin{subfigure}{\linewidth}
    \centering
    \includegraphics[width=\linewidth]{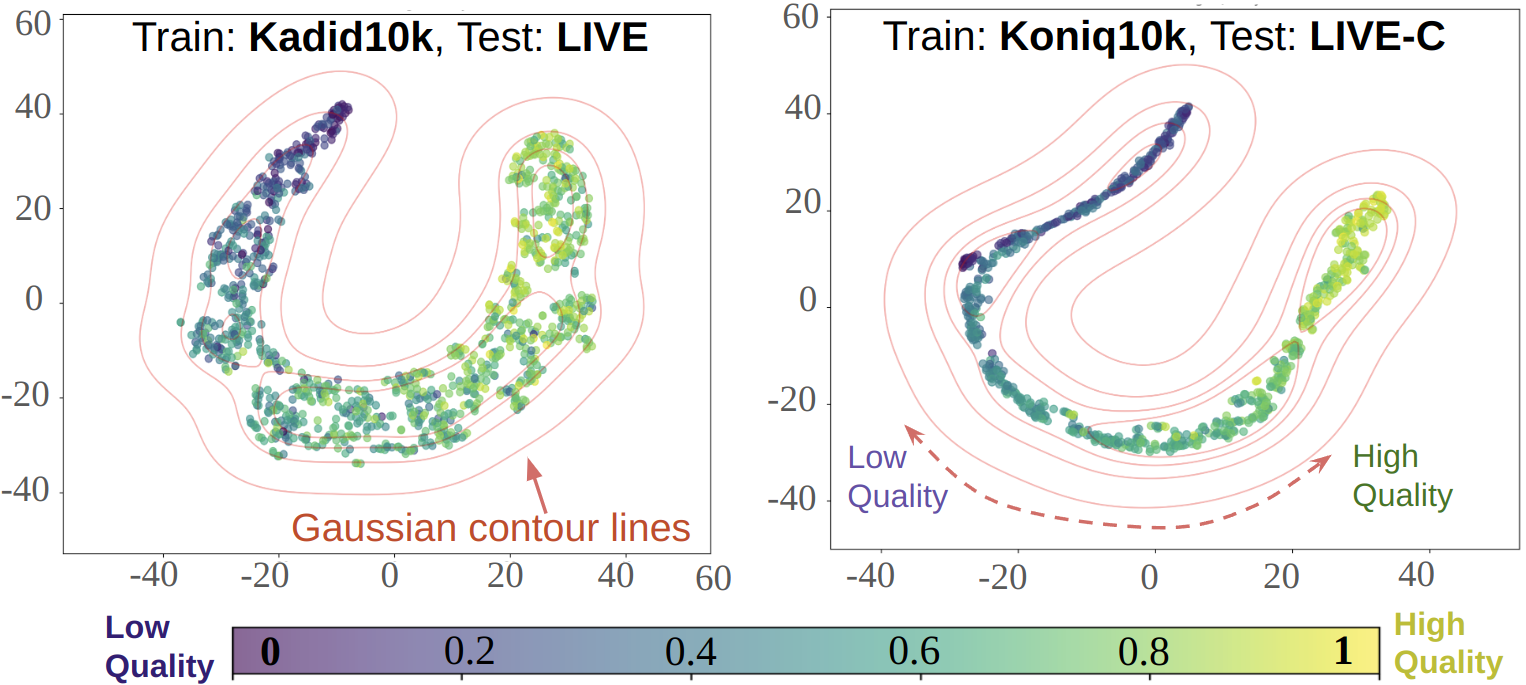}
    \caption{Cross-dataset validation across synthetic and authentic datasets.}
    \label{fig:tsne_bottom}
\end{subfigure}%
\vspace{-2mm}
\caption{Feature projections of the final DGIQA layer with t-SNE visualizations (color bar indicates normalized MOS scores).}
%\vspace{-2mm}
\label{fig:tsne}
\end{figure}

\vspace{-1mm}
We now investigate the feature distillation capabilities as well as the separability and overlap of DGIQA-predicted IQA scores. 
We use t-distributed stochastic neighbor embedding (t-SNE~\cite{van2008visualizing}) to visualize the high-dimensional feature representations of DGIQA in a 2D space. In Fig.~\ref{fig:tsne_top}, we visually inspect the clustering patterns related to different image quality levels for two large datasets: Koniq10k and Kadid10k. As seen, both visualizations reveal characteristic \textit{horseshoe patterns}, with distinct separation and smooth progression in quality levels highlighting that DGIQA effectively learns robust and transferable quality-aware features. The same pattern is visible for cross-validation results on LIVE and LIVE-C, as shown in Fig.~\ref{fig:tsne_bottom}. The \textbf{horseshoe contours are more acute} in the cross-dataset visualization,  indicating that DGIQA can learn discriminative features that capture nuanced differences between image quality in unseen distortion types.

%representations for models trained and tested on the Koniq10k and Kadid10k datasets, respectively. The figure clearly demonstrates a distinct separation between high and low-quality images. Furthermore, Figure \ref{fig:tsne_bottom} depicts the feature distributions in cross-dataset evaluation scenarios. The left panel of Figure \ref{fig:tsne_bottom} displays features extracted from the LIVE dataset using a model trained on Kadid10k, while the right panel shows features from LIVE-C, obtained using a model trained on Koniq10k. 

% \begin{figure*}[htbp]
% \centering
% \includegraphics[width=0.49\linewidth]{imgs/D2G_graph.png}
% \includegraphics[width=0.49\linewidth]{imgs/Flare7k_graph.png}
% \centering
% \includegraphics[width=0.49\linewidth]{imgs/IHAZE_graph.png}
% \includegraphics[width=0.49\linewidth]{imgs/GOPRO_graph.png}
% \vspace{-2mm}
% \caption{Comparison of model predictions for high-quality and low-quality images across three natural distortion datasets: D2G, Flare7k, IHAZE and GOPRO. In each plot, solid lines represent the score distributions for high-quality images, and dotted lines represent those for low-quality images. Blue curves correspond to MANIQA, red to ARNIQA, and green to DGIQA. The overlap area between high and low quality distributions for each model is shown, color-coded to match the respective model.}
% %\JI{is this figure ready? (b) and (d) seems same}
% \label{fig:separation}
% \end{figure*}

\begin{figure*}[htbp]
\centering
\includegraphics[width=0.5\linewidth]{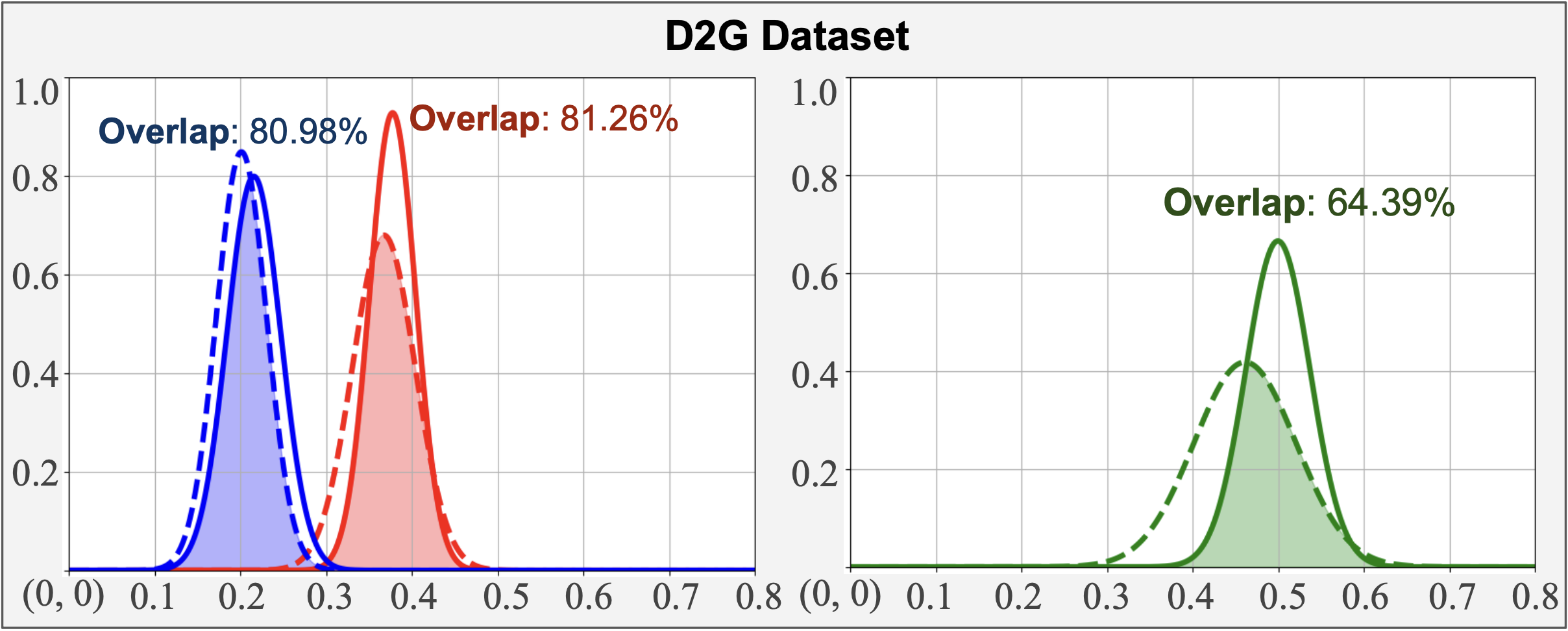}%
~
\includegraphics[width=0.5\linewidth]{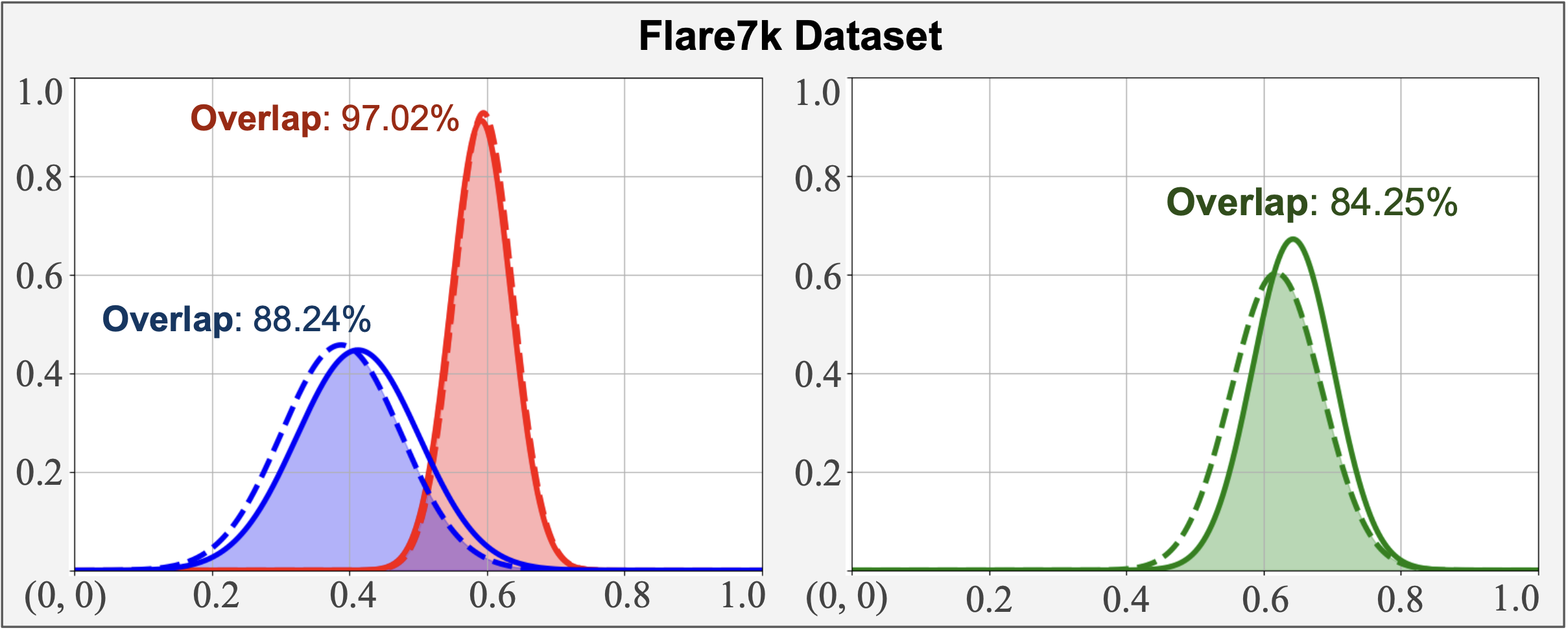} \\   \includegraphics[width=0.5\linewidth]{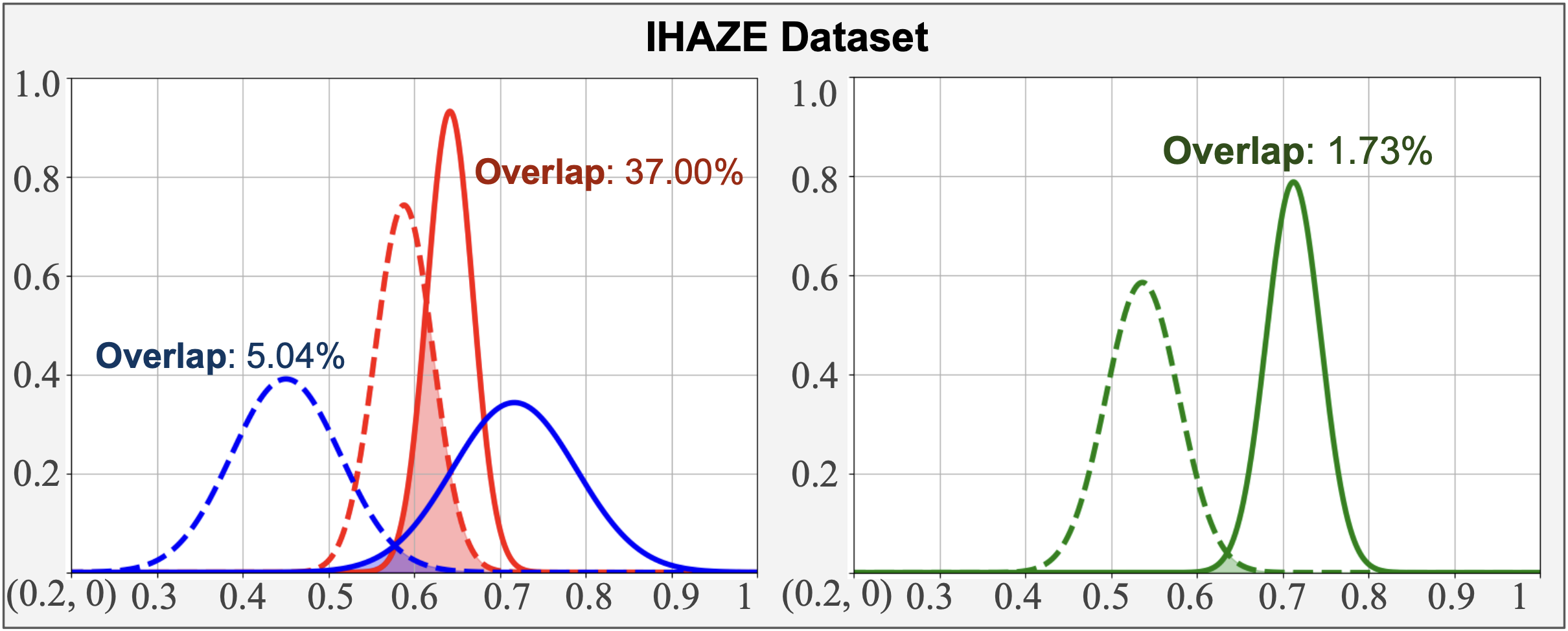}%~     
~
\includegraphics[width=0.5\linewidth]{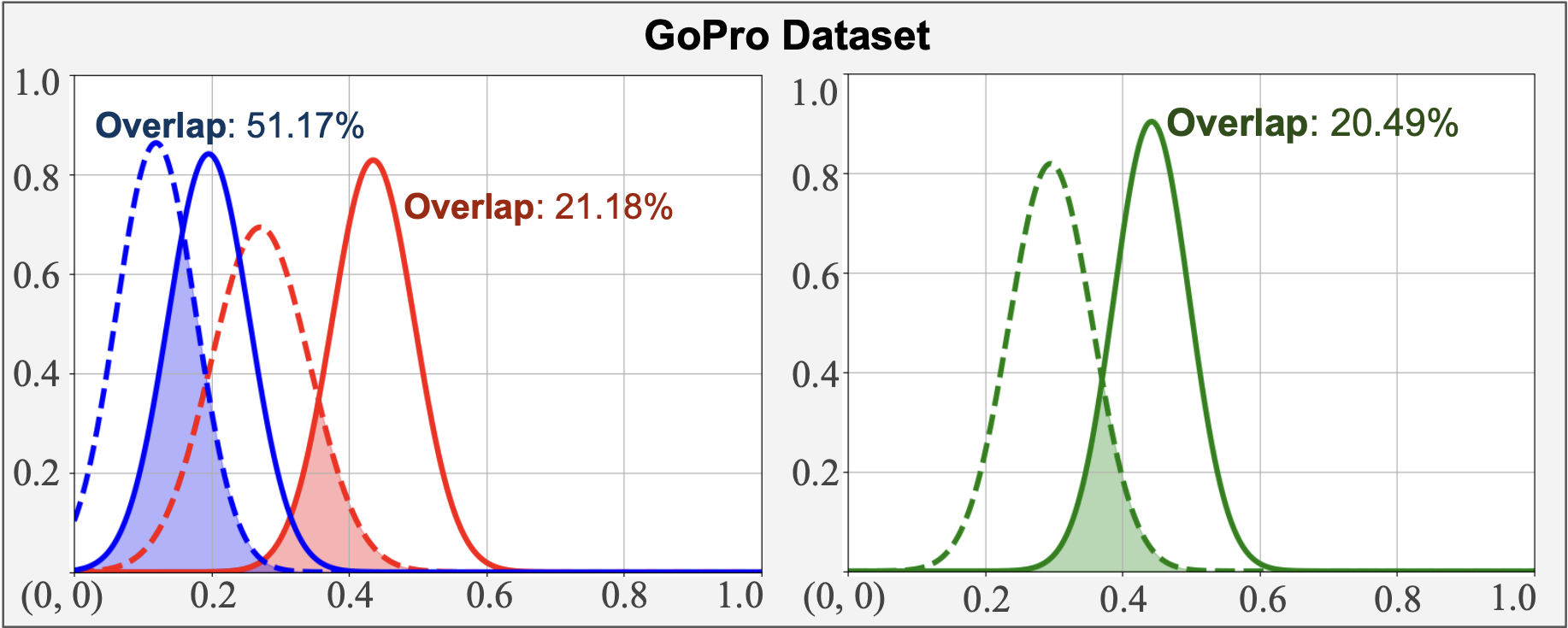} \\
\includegraphics[width=0.48\linewidth]{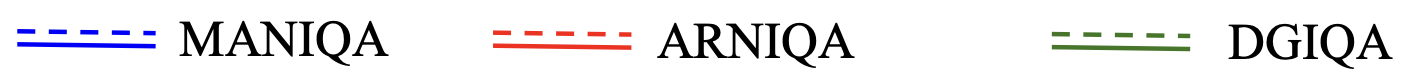}~
\hspace{3mm}
\includegraphics[width=0.48\linewidth]{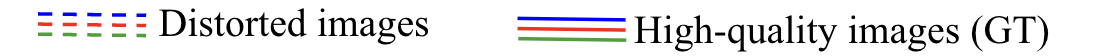}
%\vspace{-2mm}
\caption{Density separation of predicted IQA scores by DGIQA, SOTA transformer model (MANIQA), and CNN-based model (ARNIQA) are shown for four natural image distortion datasets: D2G, Flare7k, IHAZE, and GOPRO; results for LOL data is in Fig.~\ref{fig2}. Significantly lower overlap between high-quality and distorted images demonstrates a superior generalization performance of DGIQA.}
\label{fig:separation}
%\vspace{-3mm}
\end{figure*}

\subsection{Performance on Natural Image Distortions}
\vspace{-1mm}
To further demonstrate our model's generalization capabilities, we evaluated it on five datasets containing a diverse collection of natural distortions such as low light effects, lens flare, haze, and motion blur. We make use of benchmark datasets in other imaging domains, specifically: (\textbf{i}) LOL~\cite{Chen2018Retinex} and D2G~\cite{KHAN2021115034} datasets consisting of well-lit and low-light images with varying quality;  
(\textbf{ii}) Flare7k~\cite{dai2022flare7k} dataset containing images with artificially added lens flare; (\textbf{iii}) IHAZE~\cite{I-HAZE_2018} dataset with high-quality (clear) images paired with corresponding low-quality (hazy) images; and (\textbf{iv}) GoPro~\cite{nah2017deep} dataset having high-quality images paired with low-quality images distorted by motion blur.

For these experiments, we train DGIQA and a few prominent SOTA models on the Koniq10k dataset. For evaluation, we compute scores for both high-quality (ground truth) and low-quality (distorted) images and represent them as separate Gaussian distributions. Our intuition is that a well-generalized model would produce distinct distributions with less overlap between the two distributions. To this end, we propose this `{density separation}' as a \textbf{new evaluation criterion} for NR-IQA generalization performance. Fig.~\ref{fig:separation} illustrates the results for DGIQA and two SOTA models: MANIQA~\cite{yang2022maniqa} and ARNIQA~\cite{agnolucci2024arniqa} as baselines.

\begin{figure}[h]
    \vspace{-1mm}
    \centering
    \includegraphics[width=\linewidth]{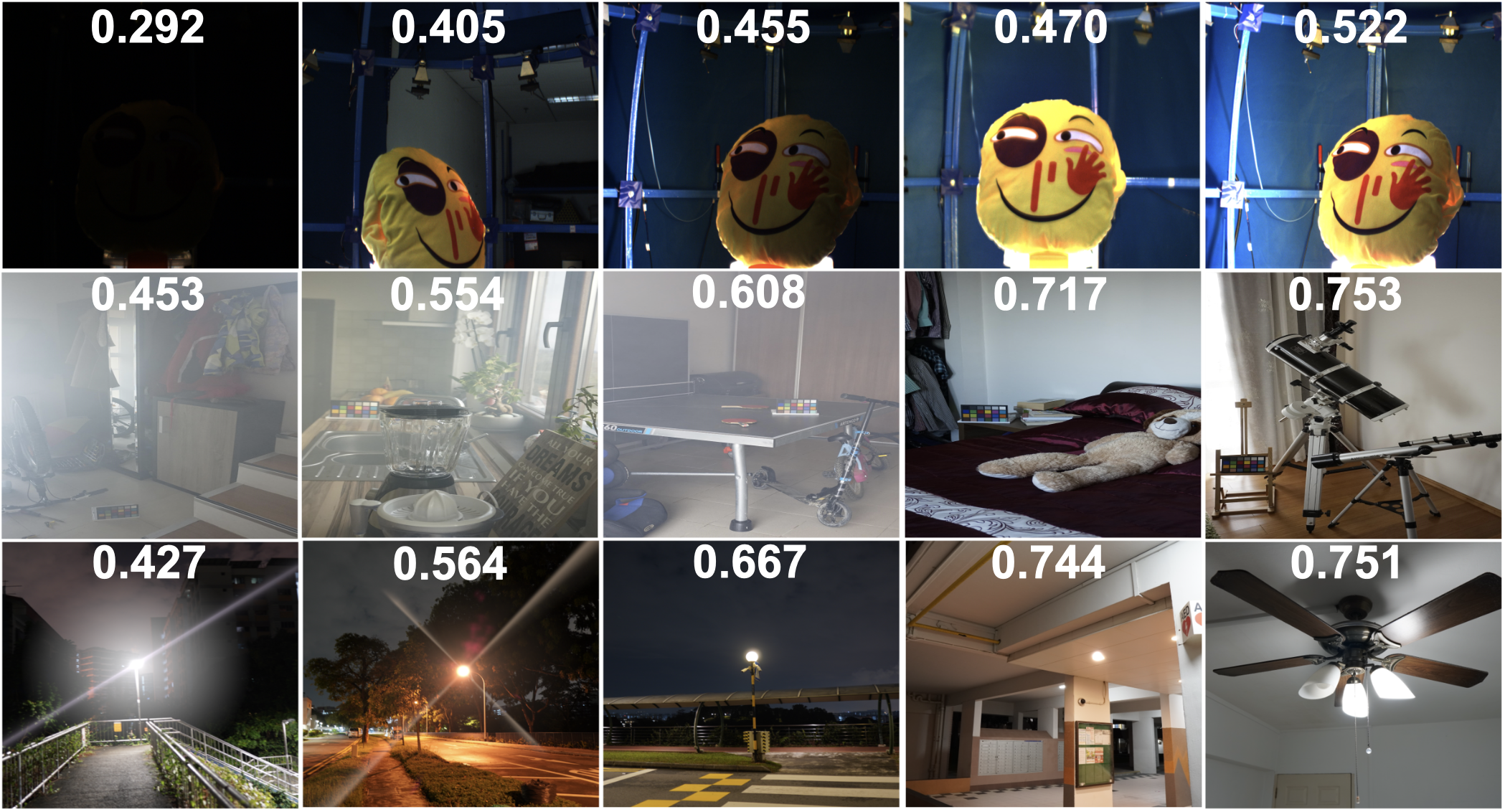}%
    \vspace{-2mm}
    \caption{Consistent IQA predictions of DGIQA are shown for: (top row) low-light effects, (middle) haze, and (bottom row) lens flares -- with decreasing levels of distortions from left to right.}%
    \label{fig:levels}
    \vspace{-1mm}
\end{figure}

\vspace{1mm}
\noindent
\textbf{Analyses and Discussion.} As shown in Fig.~\ref{fig:separation}, DGIQA achieves significantly better density separation across all datasets. Compared to the SOTA baselines, our model demonstrates $41$-$50$\%, $20$-$21$\%,  $5$-$13$\%, $65$-$95$\%, and $3$-$60$\% less overlaps in LOL, D2G,  Flare7k, IHAZE, and GOPRO datasets, respectively. Note that the subtle differences caused by lens flare make Flare7k the most challenging dataset, where all models experience significant overlap. Nevertheless, DGIQA still outperforms the baseline performance by considerable margins. It showcases the strongest generalization performance on the IHAZE dataset, achieving only $1.73$\% overlap. We hypothesize that although hazy effects degrade RGB information, depth features remain relatively stable, enabling DGIQA to perceive structural cues for more discriminative IQA learning. The consistent generalization performance of our model is more apparent from the qualitative results in Fig.~\ref{fig:levels} and Appendix~\ref{sec:Appendix-AdditionalExp}.

\vspace{1mm}
\noindent
\textbf{Limitations and Challenges.} While DGIQA demonstrates strong performances across various IQA tasks, it faces limitations in scenarios where scene depth is too noisy or not informative. For images with no significant depth variation (\eg, on uniformly textured surfaces, flat landscapes) or images captured from a single focal plane, the depth guidance does not improve IQA performance. In some cases, Depth-CAR fails to project useful features, causing model divergence. Another challenge is that strong backbone feature extractors increase the computational demands, limiting off-the-shelf use in real-time applications. Nevertheless, model compression (quantization, low-rank factorization) operations can generate \textit{light versions} to balance the robustness-efficiency trade-off for real-time use. 

%model struggle to generalize well across different datasets, where models trained on a particular dataset may learn specific features that are not fully representative of the distortions or content in other datasets. % depth noisy cannot be the same issue twice. 
%depth data can sometimes be noisy or inconsistent, especially in challenging conditions such as low light or occlusions. 

\vspace{1mm}
\noindent
\textbf{Additional Results}. Supplementary results, ablation studies, and analyses are in the \underline{Appendix section}. Moreover, DGIQA model and inference code \underline{are available at:} \url{https://github.com/uf-robopi/DGIQA}

\section{Conclusion}
\vspace{-2mm}
In this paper, we introduce DGIQA, a novel approach to NR-IQA that integrates Transformers' patch-level embeddings with CNN hierarchical features through Transformer-CNN Bridges (TCBs) for comprehensive feature distillation. We also incorporate multimodal IQA learning with depth-guided cross-attention and refinement (Depth-CAR) to focus on salient objects and structural details, enhancing IQA performance. The projection functions in TCBs reduce model parameters by $23.3$\%, while still achieving SOTA performance. DGIQA demonstrates strong generalization, surpassing SOTA models in cross-dataset validations and assessing unseen natural distortions. Future works will explore a computationally light version of DGIQA for real-time video content moderation and filtering applications.

\section*{Acknowledgment}
This work is supported in part by the National Science Foundation (NSF) award \#$2330416$ and the University of Florida (UF) research grant \#$132763$. 

%\input{rebuttal}
%%%%%%%% REFERENCES
{\small
\bibliographystyle{ieee_fullname}
\bibliography{refs}
}

\newpage

% \begin{itemize}
%     \item \url{https://openaccess.thecvf.com/content/CVPR2023/supplemental/Saha_Re-IQA_Unsupervised_Learning_CVPR_2023_supplemental.pdf}
%     \item \url{https://openaccess.thecvf.com/content_cvpr_2018/Supplemental/1335-supp.pdf}
%     \item find a few more
% \end{itemize}

\begin{comment}

\begin{figure*}[ht]
    \centering
    \label{t-sne}
    \begin{subfigure}[t]{\columnwidth}
        \centering
        \includegraphics[width=\textwidth]{imgs/tsne_live_kadid.png}  % Replace with your image file
        \caption{Test on LIVE dataset with pretrained Kadid10k weights}
    \end{subfigure}
    \hfill
    \begin{subfigure}[t]{\columnwidth}
        \centering
        \includegraphics[width=\textwidth]{imgs/tsne_live_koniq.png}  % Replace with your image file
        \caption{Test on LIVE dataset with pretrained Koniq10k weights}
    \end{subfigure}
    \vskip\baselineskip
    \begin{subfigure}[t]{\columnwidth}
        \centering
        \includegraphics[width=\textwidth]{imgs/tsne_livec_kadid.png}  % Replace with your image file
        \caption{Test on LIVE Challenge dataset with pretrained Kadid10k weights}
    \end{subfigure}
    \hfill
    \begin{subfigure}[t]{\columnwidth}
        \centering
        \includegraphics[width=\textwidth]{imgs/tsne_livec_koniq.png}  % Replace with your image file
        \caption{Test on LIVE Challenge dataset with pretrained Koniq10k weights.}
    \end{subfigure}
    \caption{t-SNE representations of feature embeddings before the final fully connected layer for different training and testing configurations. Dots indicates image opinion scores in range [0,1], light green is 1 and dark purple is 0.}
    
\end{figure*}
\end{comment}
\appendix
\section*{Appendix}
% \section{Introduction}
% \vspace{-1mm}
% %rewrite
% In this supplementary material, we provide additional results and an in-depth analysis of our DGIQA model. Specifically, we present: (\textbf{i}) Hyperparameter tuning analysis on the impact of critical parameters such as the consistency loss weight $\lambda$ and the role of depth information; (\textbf{ii}) Benchmark dataset samples from LIVE~\cite{sheikh2006statistical} and LIVE-FB~\cite{ying2020patches}, showcasing authentic and synthetic distortions alongside their depth maps to highlight our multimodal input representation; and (\textbf{iii}) Additional visualizations, including t-SNE plots of feature embeddings from benchmark datasets and cross-dataset evaluations, as well as examples of final prediction results on natural image distortion datasets. 
% These analyses and results collectively demonstrate the robustness and generalization capabilities of our model across diverse scenarios.

%\vspace{-1mm}
\section{Hyperparameter Tuning}
\label{sec:Appendix-HyperparameterTuning}
\subsection{Number of Crops}
Following up on our discussion in Sec. 4.4 (Ablation Experiments) of the paper, we further analyze the impact of the number of crops on evaluation time and performance. Figure~\ref{fig:crops} illustrates the effect of varying the number of crops on Spearman’s rank order correlation coefficient (SROCC)~\cite{wang2004image}, Pearson’s linear correlation coefficient (PLCC)~\cite{zhai2020perceptual}, and validation time, with model trained on Kadid10k and tested on LIVE dataset. Our results indicate that fewer crops significantly reduce validation time but yield lower SROCC/PLCC scores. As the number of crops increases, performance stabilizes around $25$ crops with minimal fluctuations in scores, while validation time continues to rise. Thus, selecting $25$ crops achieves consistently high performance on both metrics while maintaining a reasonable validation time, striking a practical balance between computational efficiency and prediction accuracy.
% Specifically, we identify $25$ crops as the optimal trade-off between accuracy and validation efficiency. 
%\JI{which dataset and how it was evaluated? write details}

\begin{figure}[h]
    \centering
\includegraphics[width=\columnwidth]{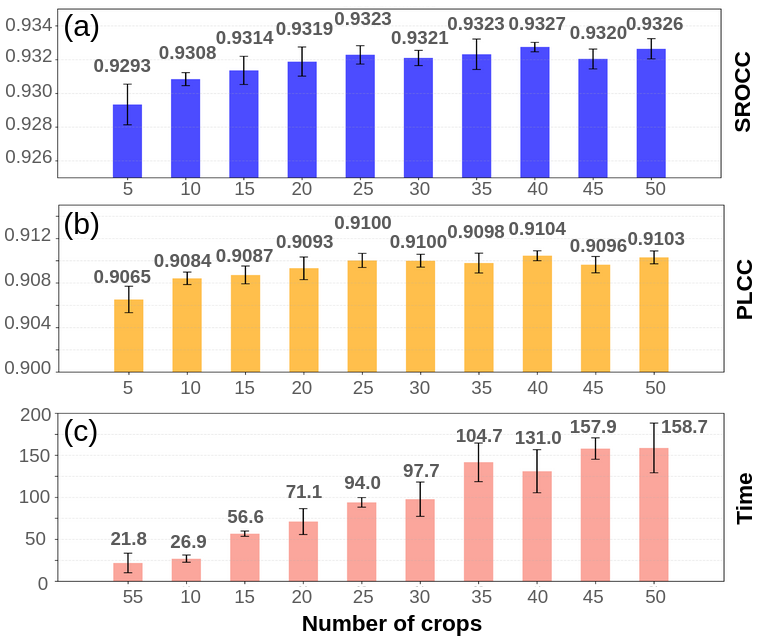}%
    \vspace{-2mm}
    \caption{Cross-dataset ablation based on different numbers of crops, trained on Kadid10k and tested on LIVE dataset, are shown for (a) SROCC; (b) PLCC; (c) Validation time in milliseconds.}
    \label{fig:crops_supplementary}
    \vspace{-2mm}
\end{figure}

\begin{figure}[h]
    \centering
    %\vspace{-2mm}
    \includegraphics[width=\columnwidth]{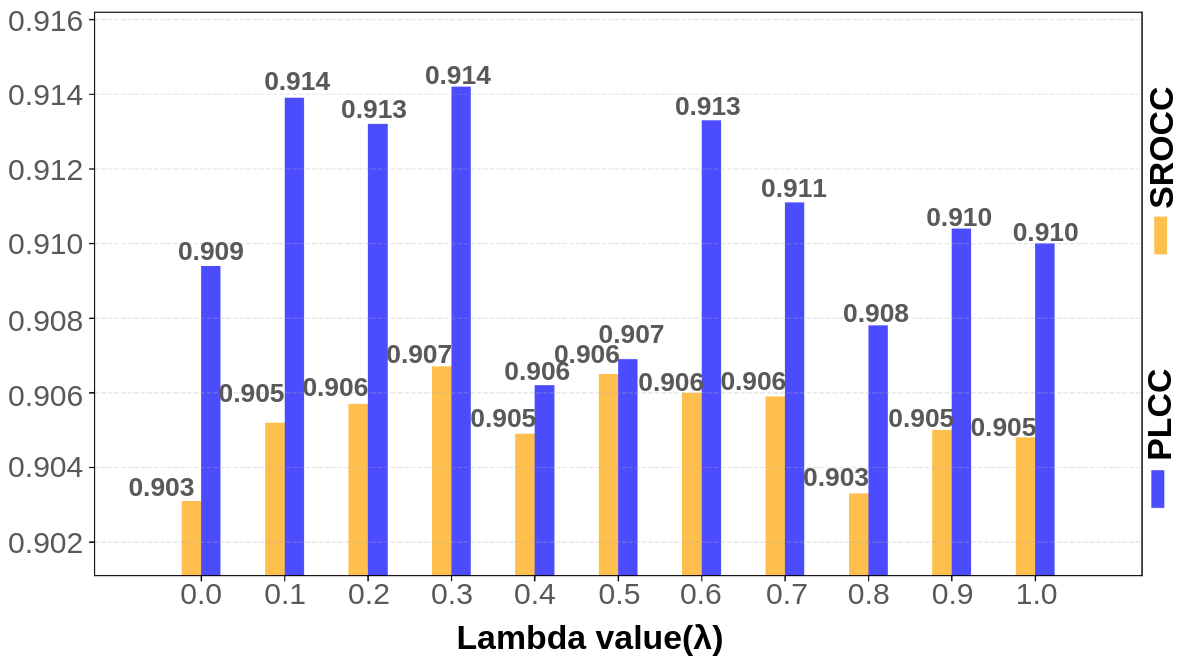}%
    \vspace{-2mm}
    \caption{DGIQA performance on LIVE-C dataset for different consistency weights (of $\lambda$).}
    \label{fig:lambda}
    \vspace{-3mm}
\end{figure}

\subsection{Contribution of the Consistency Loss}
The objective function employed in our model, detailed in Section 3.6 (Loss Functions) of the paper, is defined as:
\begin{equation} \mathcal{L}_{\text{Total}} = \mathcal{L}_{\text{SME}} + \lambda \,\mathcal{L}_{\text{CL}} \end{equation}
where $\lambda$ is a hyperparameter that balances the contribution of the consistency loss, $\mathcal{L}_{\text{CL}}$. 
\begin{figure*}[ht]
    \centering
    \includegraphics[width=0.8\linewidth]{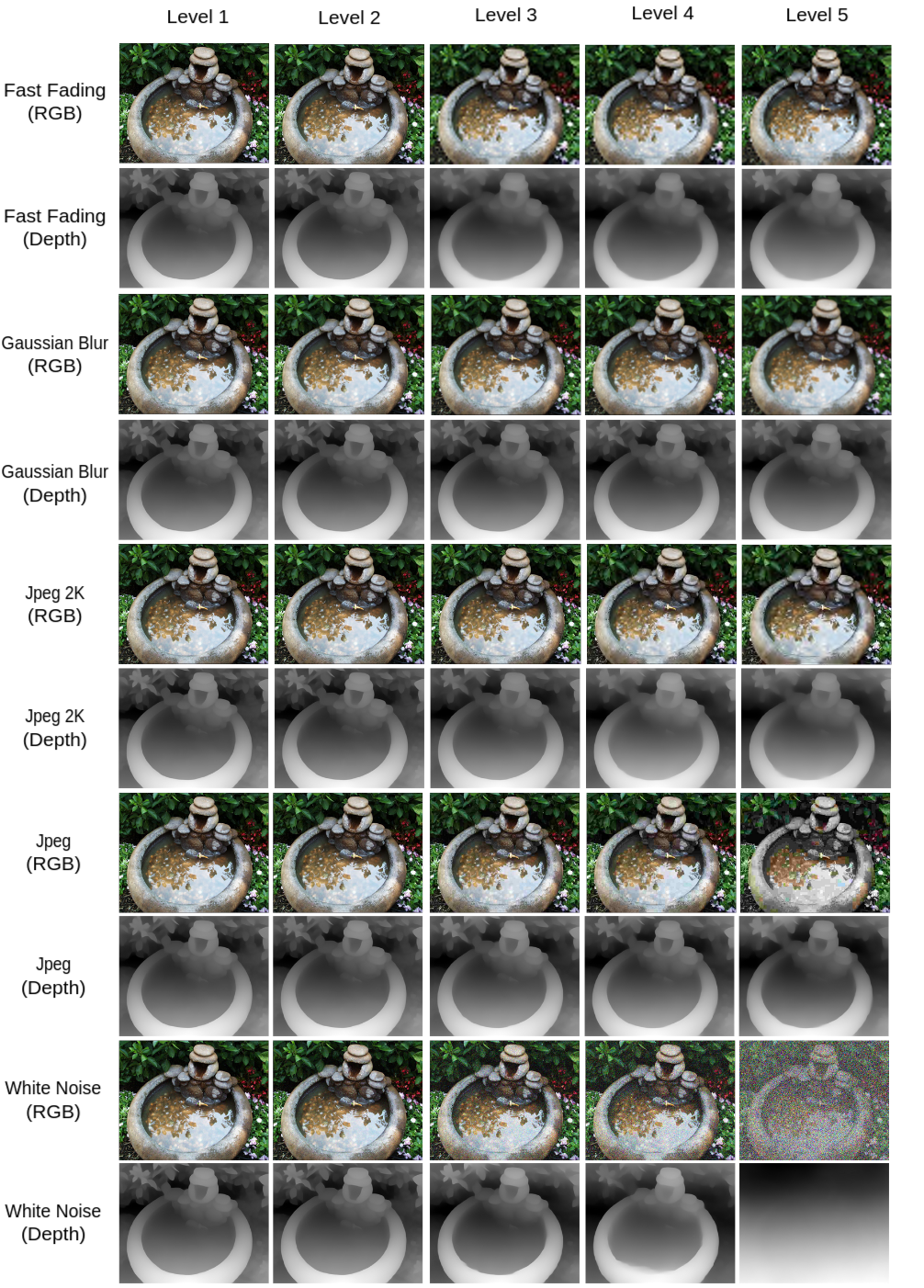}%
    \vspace{-2mm}
    \caption{LIVE distortion images and their depth map samples. LIVE dataset includes 5 distortion types: {\tt Fast Fading}, {\tt Gaussian Blur}, {\tt JPEG 2K}, {\tt JPEG}, and {\tt White Noise}, each having a distortion level from 1 (minimum distortion) to 5 (maximum distortion).}
    \label{fig:LIVE RGBD}
    \vspace{-3mm}
\end{figure*}

\begin{figure*}[ht]
    \centering
    \includegraphics[width=\linewidth]{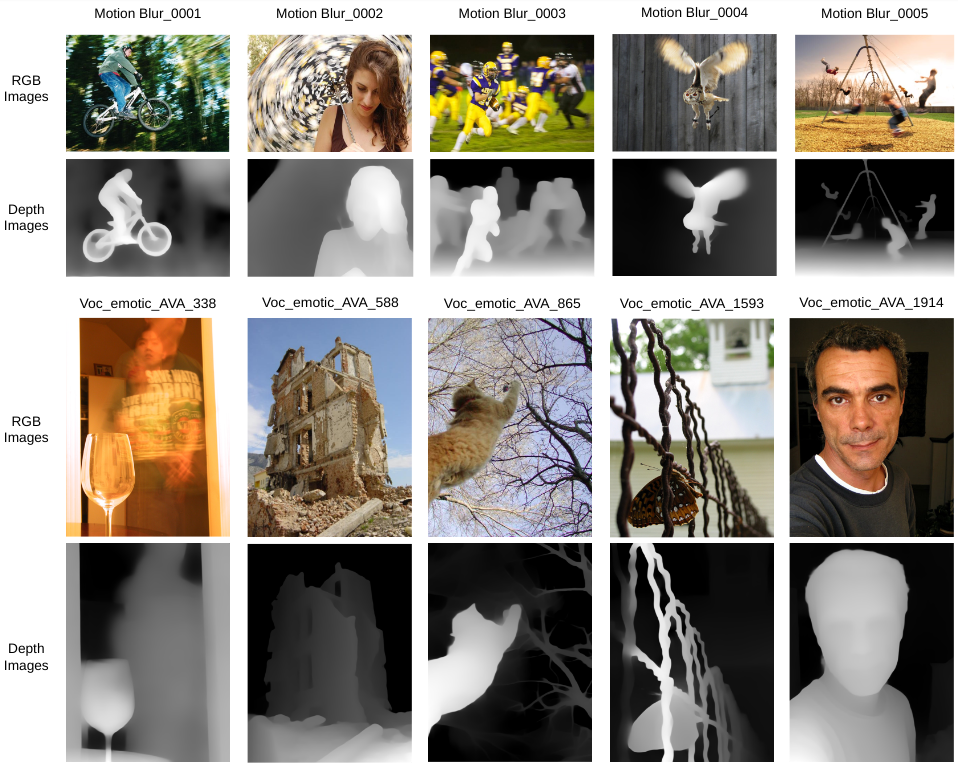}%
    \vspace{-1mm}
    \caption{A few samples from the LIVE-FB synthetic distortion images and their depth maps are shown. Images are chosen from two inherent categories: {\tt Motion Blur} and {\tt Voc\_emotic\_AVA}. }
    \label{fig:LIVEFB RGBD}
    \vspace{-2mm}
\end{figure*}

Figure~\ref{fig:lambda} illustrates the impact of varying $\lambda$ on performance metrics for training and validation on the LIVE-C~\cite{ghadiyaram2015massive} dataset for one training seed, focusing on SROCC and PLCC. The results indicate that the SROCC score reaches its minimum at $\lambda = 0$, while PLCC reaches its minimum at $\lambda = 0.4$. Both metrics achieve their maximum values when $\lambda = 0.3$. Due to image clarity constraints, we report results rounded to three decimal places; however, detailed trends can be observed from the bar chart. Overall, the results in Figure~\ref{fig:lambda} demonstrate that setting $\lambda = 0.3$ allows the model to achieve optimal performance by maximizing both correlation metrics.%\JI{need more detailed analyses}

% Waiting for results for DGIQA will update them afternoon

\vspace{-2mm}
\section{Distortion and Depth Images}
\vspace{-2mm}
Our proposed DGIQA model leverages distorted RGB images and their corresponding depth maps to enhance feature representation, facilitating improved IQA learning. To make the training process more effective and streamlined, we pre-compute all depth maps using DepthAnything~\cite{yang2024depth} when ground truth depth maps are unavailable. Figure~\ref{fig:LIVE RGBD} illustrates examples from the LIVE~\cite{sheikh2006statistical} dataset, showcasing images subjected to five synthetic distortions: {\tt Fast Fading}, {\tt Gaussian Blur}, {\tt JPEG 2000}, {\tt JPEG}, and {\tt White Noise}. Each row highlights how the corresponding depth maps effectively capture structural and geometric cues despite the varying levels of distortion (Levels 1-5). Depth maps for severe distortions (\eg, White Noise Level 5) clearly demonstrate degraded structural information, reflecting the impact of heavy noise on feature extraction.

Figure~\ref{fig:LIVEFB RGBD} provides examples from the LIVEFB~\cite{ying2020patches} dataset, which contains two inherent categories of distortions: {\tt Motion Blur} and {\tt Voc\_emotic\_AVA}. The depth maps generated by DepthAnything reveal their ability to emphasize foreground regions, or \textit{salient} objects, critical for human visual attention~\cite{islam2022svam}. For instance, the {\tt Motion Blur} examples (top rows) retain the structural outlines of dynamic objects, such as bicycles and people in motion, even when RGB images exhibit heavy blurring. Similarly, the {\tt Voc\_emotic\_AVA} examples (bottom rows) showcase depth maps that capture distinct foreground elements, such as architectural structures, animals, or human figures, while providing a clear separation from the background.

Notably, these depth maps consistently highlight salient regions across both synthetic and authentic distortions, demonstrating their robustness and utility in guiding the model to focus on visually significant areas. This capability is essential for IQA, where human perception often prioritizes structural and semantic information in assessing image quality. The visualizations emphasize the effectiveness of integrating depth cues into the DGIQA learning pipeline, enabling it to handle challenging distortion scenarios with improved accuracy and robustness.

\vspace{-1mm}
\section{Additional Experiment Results}
\label{sec:Appendix-AdditionalExp}
\vspace{-1mm}
\subsection{Effective feature distillation}
\vspace{-1mm}

\begin{figure*}
    \centering
    \includegraphics[width=0.98\columnwidth]{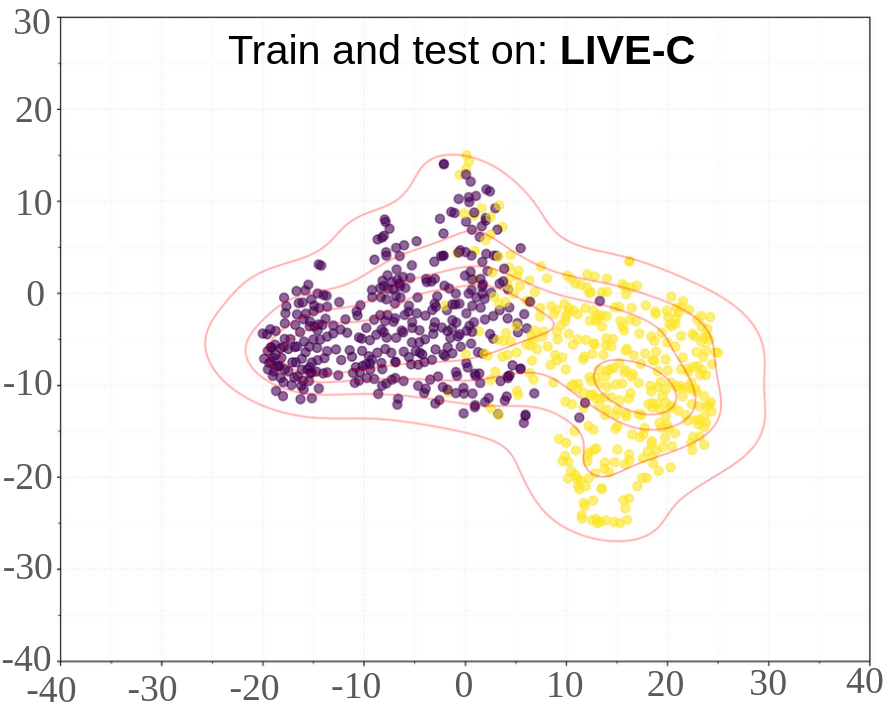}~
    \hspace{2mm}
    \includegraphics[width=0.98\columnwidth]{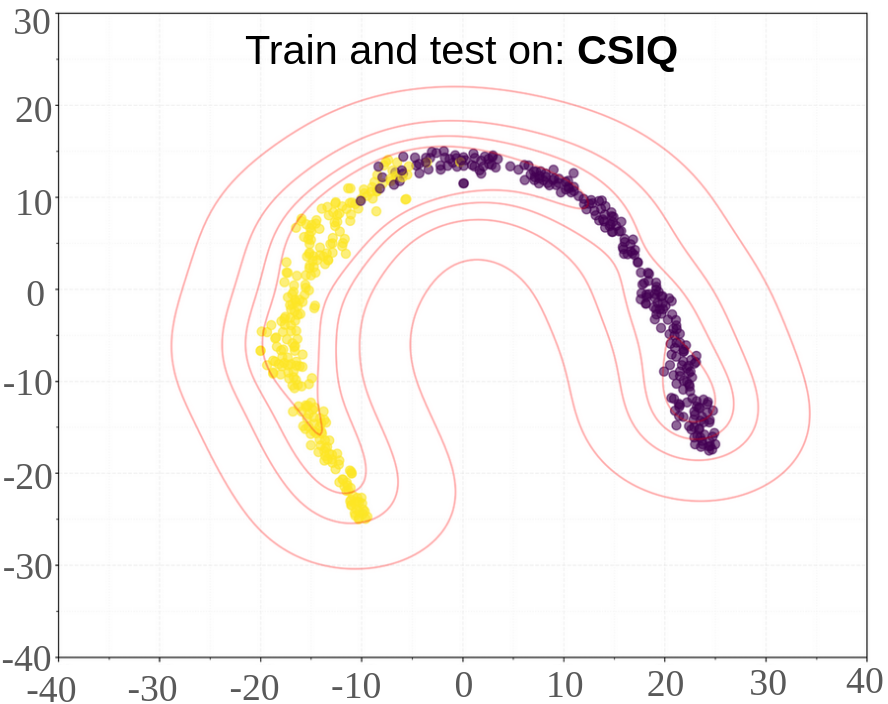} \\
    \vspace{2mm}
    \includegraphics[width=0.98\columnwidth]{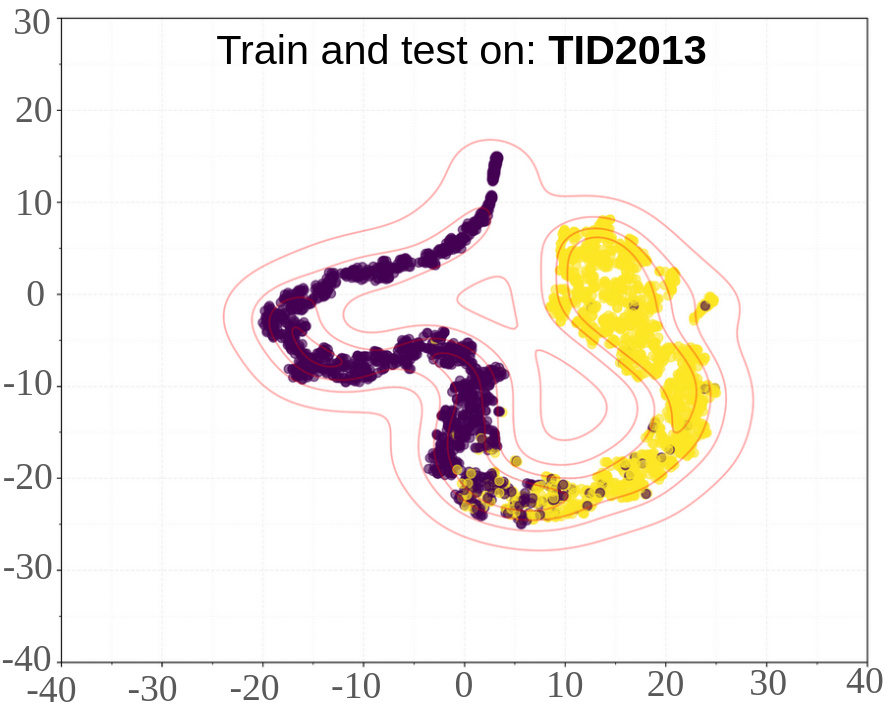}~
    \hspace{2mm}
    \includegraphics[width=0.98\columnwidth]{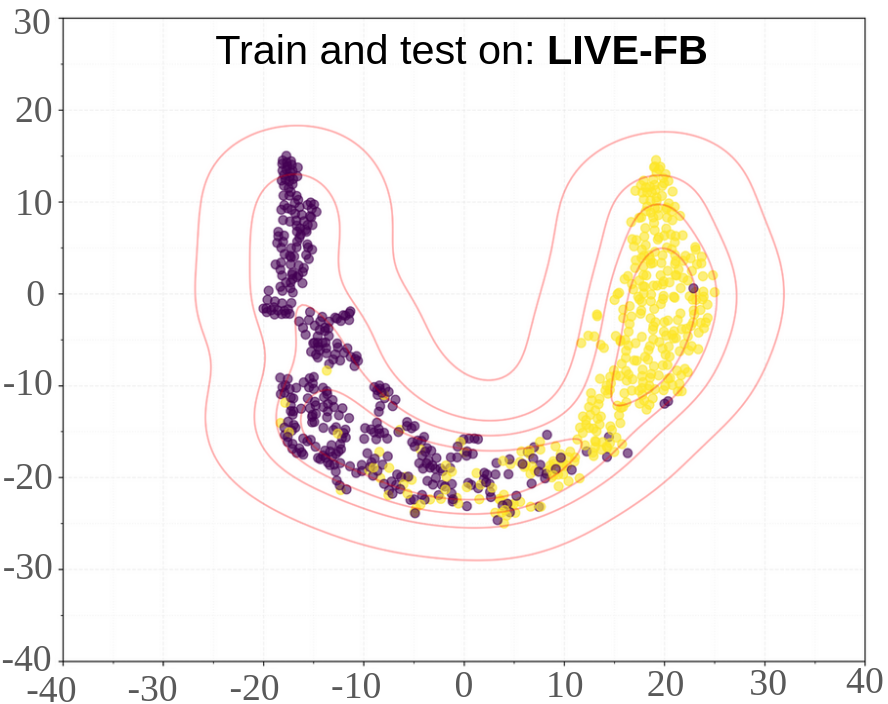} \\
    \vspace{2mm}
    \includegraphics[width=0.98\columnwidth]{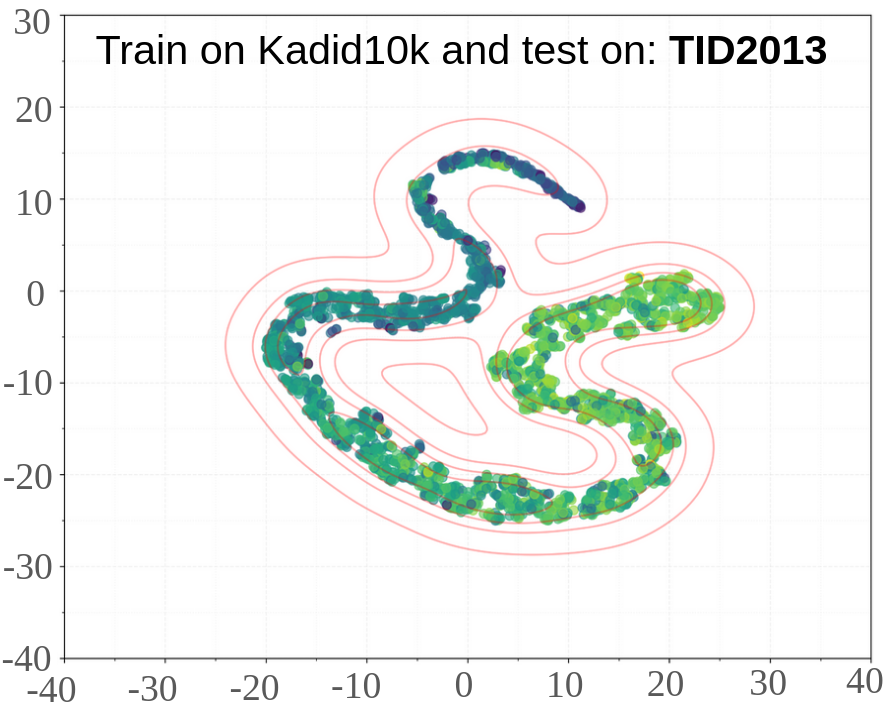}~
    \hspace{2mm}
    \includegraphics[width=0.98\columnwidth]{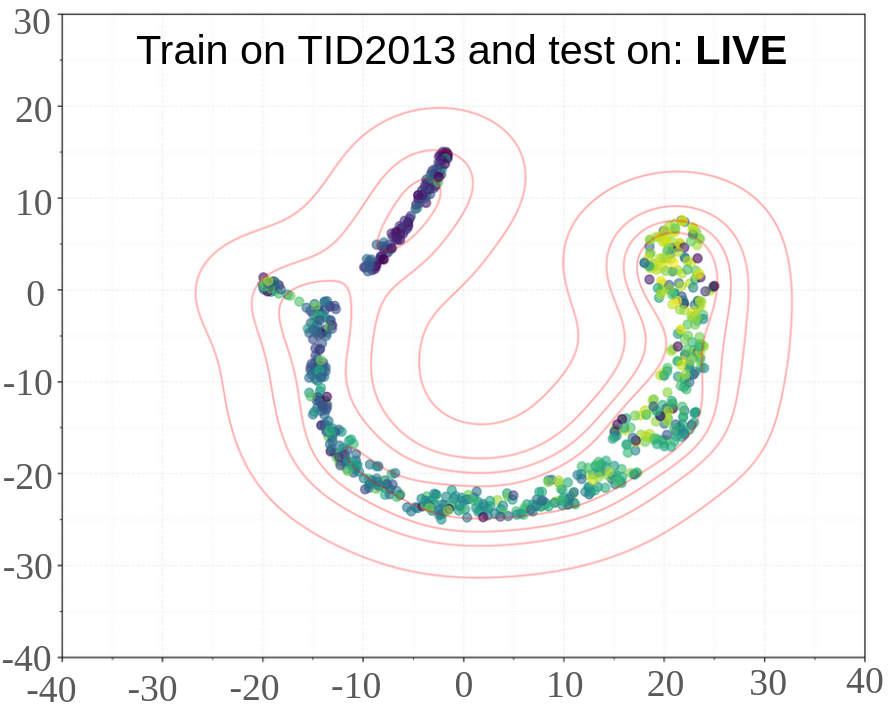} \\
    
    \includegraphics[width=0.75\textwidth]{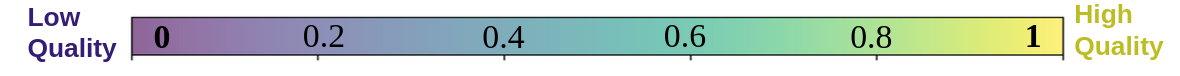}%
    \vspace{-1mm}
    \caption{t-SNE visualizations of feature projections from the final DGIQA layer. For single-dataset tests, the visualization shows separations between "high-quality" and "low-quality" images. For cross-dataset tests, the visualization uses continuous colors to represent quality scores, ranging from 0 (completely distorted) to 1 (original image).}
    \vspace{-2mm}
    \label{fig:tsne_supplementary}
\end{figure*}

We present high-dimensional feature representations using t-distributed stochastic neighbor embeddings (t-SNE). Figure~\ref{fig:tsne_supplementary} visualizes feature embeddings from single-dataset training and testing on LIVE-C~\cite{ghadiyaram2015massive}, CSIQ~\cite{larson2010most}, TID2013~\cite{ponomarenko2015image}, and LIVE-FB~\cite{ying2020patches}, as well as cross-dataset evaluations where the model is trained on Kadid10k~\cite{lin2019kadid} and tested on TID2013, and trained on TID2013 and tested on LIVE. Each plot demonstrates the model's ability to separate high-quality (yellow) and low-quality (purple) images, with the gradients representing intermediate-quality levels.

In single-dataset scenarios, t-SNE embeddings for LIVE-C, CSIQ, TID2013, and LIVE-FB reveal distinct clustering of quality levels. LIVE-C shows clear distinctions between low- and high-quality images, with some overlap in intermediate scores. CSIQ and TID2013 exhibit ``\textit{horseshoe}'' patterns indicating a smooth progression from low to high-quality embeddings. In cross-dataset evaluations, the embeddings demonstrate the model's generalization capabilities. When trained on Kadid10k and tested on TID2013, feature separation remains robust, with high- and low-quality embeddings forming distinct clusters. Similarly, when trained on TID2013 and tested on LIVE, the embeddings maintain clear separation, though slightly more compact due to domain shift. These visualizations demonstrate DGIQA's strong feature discrimination and generalization, consistently producing meaningful separations across synthetic, authentic, and cross-dataset scenarios.

\begin{table*}[t]
\centering
\caption{SROCC and PLCC results comparison of DGIQA against other IQA models on four benchmark datasets. The top two scores are highlighted in {\color{blue}blue}, with the best scores in \textbf{\color{blue}bold}.}
\vspace{-2mm}
\begin{tabular}{l||cccccccc}
\Xhline{2\arrayrulewidth}
 & \multicolumn{2}{c}{LIVE~\cite{sheikh2006statistical}} & \multicolumn{2}{c}{CSIQ~\cite{larson2010most}} & \multicolumn{2}{c}{TID2013~\cite{ponomarenko2015image}} & \multicolumn{2}{c}{Kadid10k~\cite{lin2019kadid}} \\
\cline{2-3} \cline{4-5} \cline{6-7} \cline{8-9}
\textbf{Model} & \textbf{SROCC} & \textbf{PLCC} & \textbf{SROCC} & \textbf{PLCC} & \textbf{SROCC} & \textbf{PLCC} & \textbf{SROCC} & \textbf{PLCC} \\
\Xhline{2\arrayrulewidth}
PSNR~\cite{5596999} & 0.881 & 0.868 & 0.820 & 0.824  & 0.643 & 0.675 & 0.677 & 0.680  \\
SSIM~\cite{wang2004image}  & 0.921 & 0.911 & 0.854 & 0.835 & 0.642 & 0.698 & 0.641 & 0.633  \\
FSIM~\cite{zhang2011fsim}  & 0.964 & 0.954 & 0.934 & 0.919 & 0.852 & 0.875 & 0.854 & 0.850  \\
PieAPP~\cite{8578292}  & 0.915 & 0.905 & 0.900 & 0.881 & 0.877 & 0.850 & 0.869 & 0.869 \\
LPIPS~\cite{zhang2018perceptual}  & 0.932 & 0.936 & 0.884 & 0.906 & 0.673 & 0.756 & 0.721 & 0.713  \\
Re-IQA-FR~\cite{saha2023re}  & \color{blue}0.969 & \color{blue}0.974 & 0.961 & 0.962 & \color{blue}0.920 & \color{blue}0.921 & \color{blue}0.933 & \color{blue}0.936  \\
ARNIQA-FR~\cite{agnolucci2024arniqa}  & \color{blue}0.969 & 0.972 & \color{blue}0.971 & \textbf{\color{blue}0.975} & 0.898 & 0.901 & 0.920 & 0.919  \\
\textbf{DGIQA-FR~(ours)} & \textbf{\color{blue}0.983} & \textbf{\color{blue}0.977} & \textbf{\color{blue}0.973} & \color{blue}0.963 & \textbf{\color{blue}0.934} & \textbf{\color{blue}0.933} & \textbf{\color{blue}0.943} & \textbf{\color{blue}0.946} \\
\Xhline{2\arrayrulewidth}
\end{tabular}
\label{Tab:FR-IQA results}
\vspace{-2mm}
\end{table*}

\subsection{Grad-Cam visualizations}
\begin{figure*}[ht]
    \centering
    \begin{subfigure}[t]{\textwidth}
        \centering
        \includegraphics[width=\textwidth]{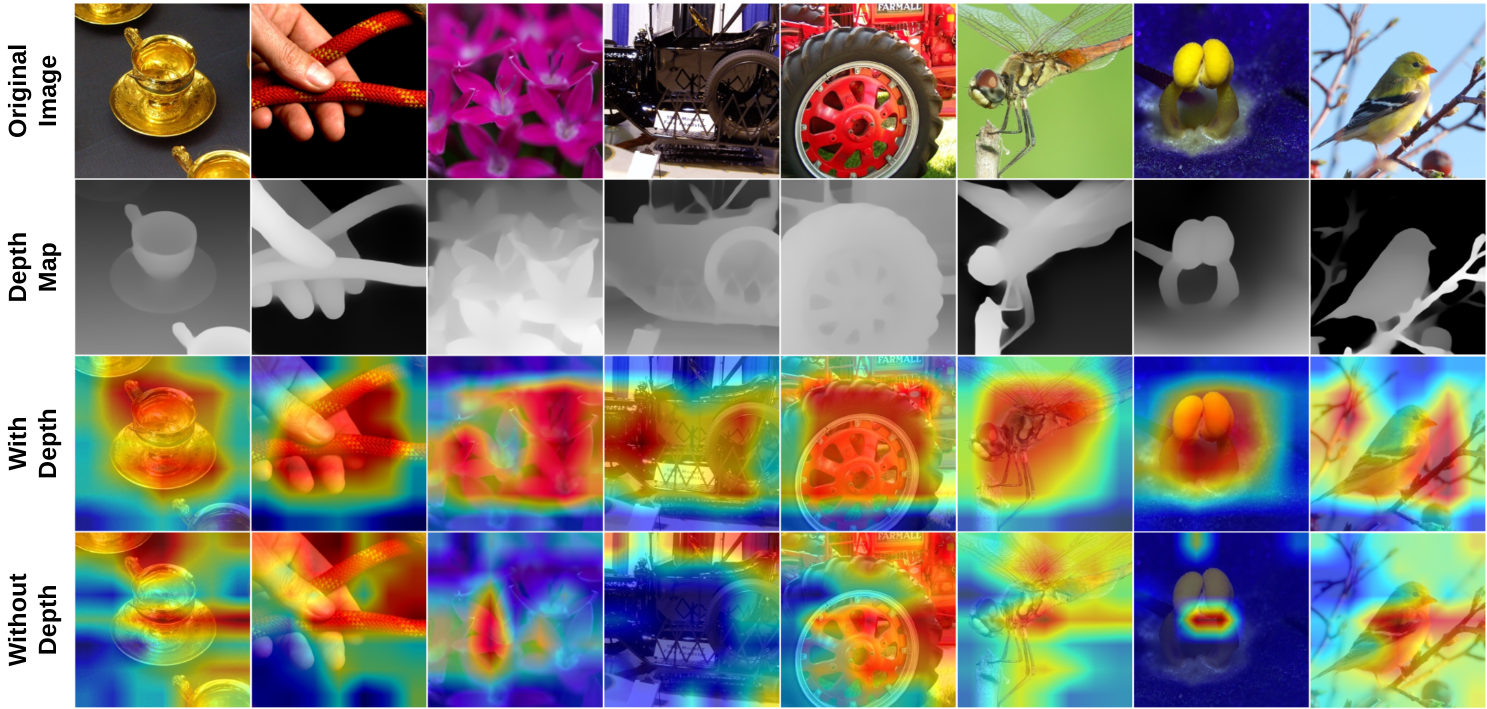}
        \caption{Close-up scenes with a focus on dominant foreground objects.}
        \label{fig:gradcam_closeup}
    \end{subfigure}
    
    \vspace{2mm}
    \begin{subfigure}[t]{\textwidth}
        \centering
        \includegraphics[width=\textwidth]{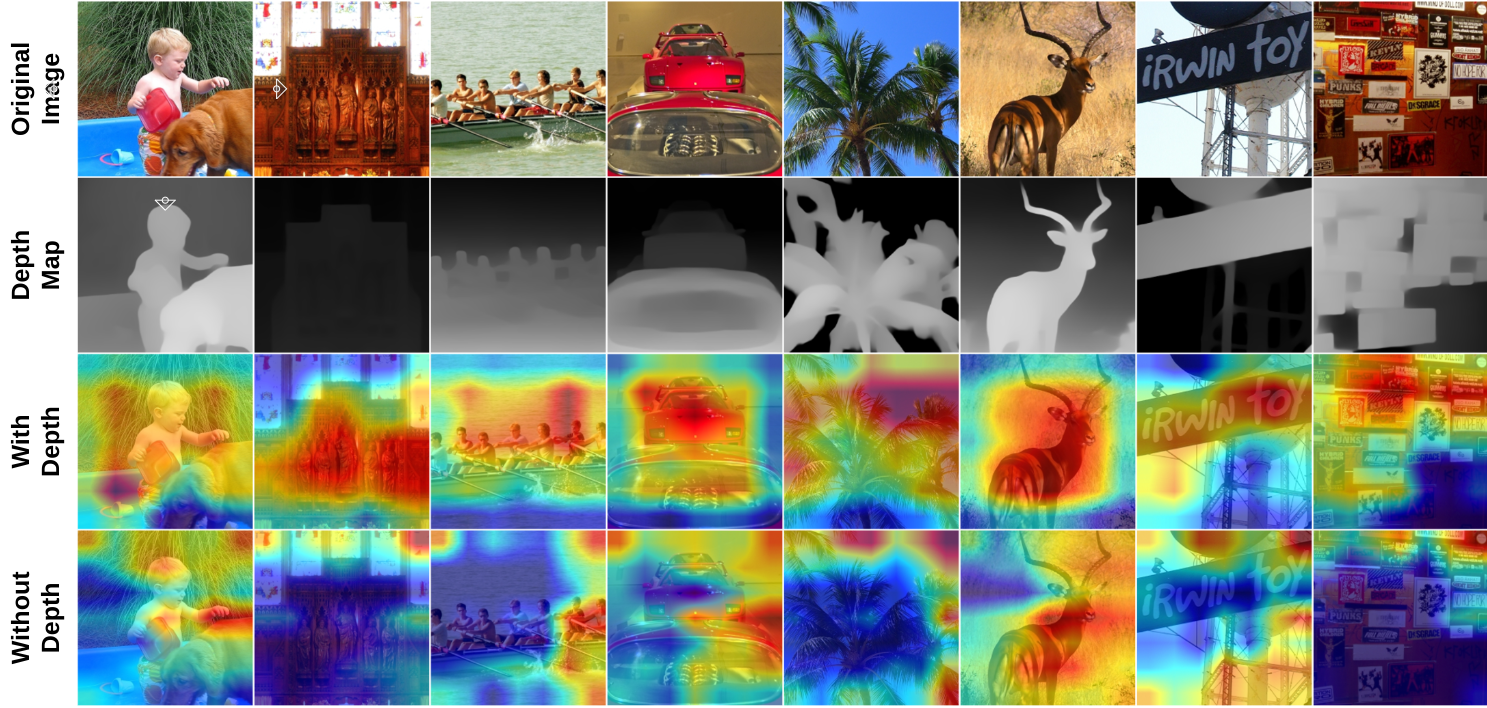}
        \caption{Scenes with distant objects and broader depth variations.}
        \label{fig:gradcam_distant}
    \end{subfigure}
    \vspace{-2mm}
    \caption{Additional examples of Grad-CAM~\cite{8237336} visualizations for the DGIQA model, highlighting the impact of depth. Each row displays the original RGB image, the Depth Map generated by DepthAnything~\cite{yang2024depth}, and the Grad-CAM outputs with and without depth input, respectively. (a) illustrates close-up scenes, while (b) features distant scenes with greater depth variations.}
    \label{fig:gradcam_suppl}
    \vspace{-2mm}
\end{figure*}

We integrated depth information into our model to enhance its focus on closer or salient objects, aligning more closely with human visual perception. Here, we present additional Grad-CAM~\cite{8237336} visualizations from the dilation convolution layer, comparing two model configurations: the full model with depth input and an ablated version without depth input (and without Depth-CAR module). Figure~\ref{fig:gradcam_suppl} presents additional examples that demonstrate how incorporating depth helps the model to focus on salient objects. These examples are categorized into two types of scenes: (\textbf{i}) scenes where the objects are positioned closer to the camera (Figure \ref{fig:gradcam_closeup}), allowing the model to easily emphasize prominent features, and (\textbf{ii}) scenes containing distant objects (Figure \ref{fig:gradcam_distant}) with more depth variations.

The visualizations in Figure~\ref{fig:gradcam_suppl} validate that the full DGIQA model consistently allocates attention to foreground regions and salient objects, while the ablated model's focus is more scattered. This consistent pattern across various scene types and distortion levels emphasizes the importance of depth information in guiding the model’s attention toward human-like perceptual evaluation.

% \begin{figure*}
%     \centering
%     \includegraphics[width=\textwidth]{suppimgs/gradcam_examples_closeup.png}
%     \\
%     \includegraphics[width=\textwidth]{suppimgs/gradcam_examples_far.png}
%     %
%     \vspace{-2mm}
%     \caption{Additional examples of Grad-CAM~\cite{8237336} visualizations for the DGIQA model, highlighting the impact of depth. Each row displays the original RGB image, the Depth Map generated by DepthAnything~\cite{yang2024depth}, and the Grad-CAM outputs with and without depth input respectively. The first set of images shows close-up shots (top), while the second set features distant scenes with more depth variations (bottom). \JI{Do a subfigure (a) close-up scene and (b) distant scene with depth variations}}
%     \label{fig:gradcam_suppl}
% \end{figure*}

\begin{figure*}
    \centering
    \includegraphics[width=\textwidth]{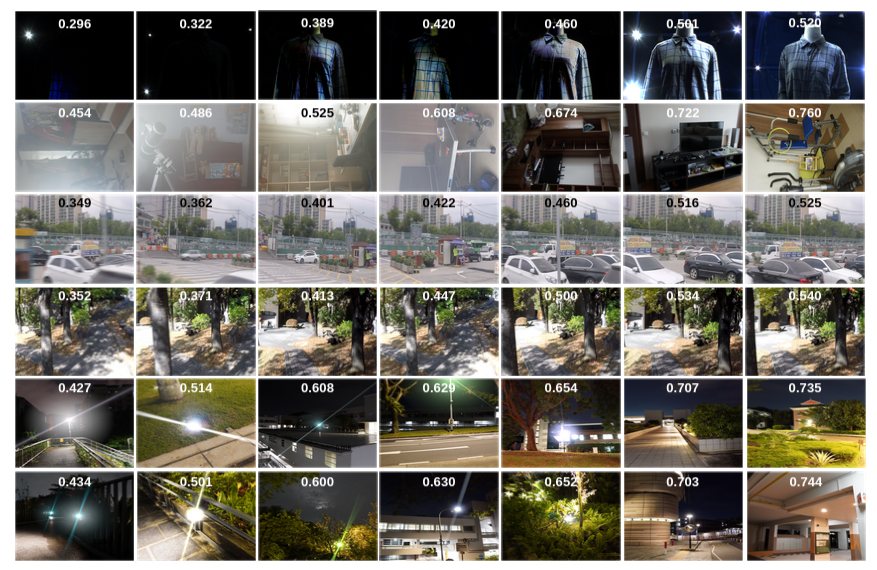}%
    \vspace{-3mm}
    \caption{IQA predictions by our DGIQA model on unseen datasets: first row shows images from D2G dataset; second row from IHAZE dataset; third and fourth rows from GoPro dataset; fifth and sixth rows from Flare7k dataset. At each column, images are arranged with decreasing levels of distortion from left to right.}%
    \label{fig:final_prediction}
    \vspace{-2mm}
\end{figure*}

\subsection{Full reference Image Quality Assessment}
To further evaluate the effectiveness of our feature extractor, we conducted experiments to adapt our model for a Full Reference Image Quality Assessment (FR-IQA) task. The model architecture remains the same as in the NR-IQA setup. In this configuration, both the distorted image and its reference image are passed through the model to obtain their respective feature maps from the dilation convolution layer. The quality score, $\hat{q}$ is then computed by taking the difference between these feature maps~\cite{saha2023re} and passing it through a fully connected (FC) layer, as follows:
\begin{equation} 
\hat{q} = \sigma \big( \mathbf{W}_{\text{fc}} \cdot (\mathbb{F}_{\text{ref}} - \mathbb{F}_{\text{dist}}) + \mathbf{b}_{\text{fc}} \big) \label{equation:FR-IQA
} 
\end{equation}
where $\mathbb{F}_{\text{ref}}$ and $\mathbb{F}_{\text{dist}}$ are the feature maps of the reference and distorted images, respectively; $\mathbf{W}_{\text{fc}}$ and $\mathbf{b}_{\text{fc}}$ are the learnable weights and biases of the FC layer; and $\sigma$ is the sigmoid function.

The training strategy for this experiment is kept similar to that of our NR-IQA learning, with the dataset split into an $80:20$ ratio for training and testing. The experiment is repeated $10$ times, and the mean SROCC and PLCC scores are computed as the final result. All other data augmentation and pre-processing steps are kept identical. We found that DGIQA-FR demonstrates faster convergence compared to DGIQA, requiring only $100$ epochs. An initial learning rate of $5e^{-5}$ is used, and a Cosine Annealing learning rate scheduler~\cite{yang2024align} is applied for smoother convergence.

Table~\ref{Tab:FR-IQA results} (see Page-4) presents the SROCC and PLCC results of our \textbf{DGIQA-FR} model, evaluated across four benchmark IQA datasets: LIVE, CSIQ, TID2013, and Kadid10k. For performance evaluation, we compare it against three closed-form metrics: PSNR~\cite{5596999}, SSIM~\cite{wang2004image}, and FSIM~\cite{zhang2011fsim}; and four learning-based metrics: PieAPP~\cite{8578292}, LPIPS~\cite{zhang2018perceptual}, REIQA-FR~\cite{saha2023re}, and ARNIQA-FR~\cite{agnolucci2024arniqa}. Among these, Re-IQA-FR and ARNIQA-FR were originally designed as NR-IQA models but were adapted for FR-IQA tasks in their respective works, serving as meaningful baselines for comparison. Without additional re-configuration or hyper-parameter tuning, our DGIQA-FR model demonstrates competitive results in both SROCC and PLCC across all four datasets. By performing comparably or better than other models, DGIQA-FR showcases its ability to adapt and generalize to FR-IQA tasks as well.

\subsection{External dataset predictions}
Figure~\ref{fig:final_prediction} showcases additional examples from natural image distortion datasets, highlighting the robustness of our DGIQA model across diverse real-world scenarios. The datasets include D2G~\cite{KHAN2021115034} for low-light conditions, IHAZE~\cite{I-HAZE_2018} for hazy degradation, GoPro~\cite{nah2017deep} for motion blur, and Flare7k~\cite{dai2022flare7k} for light flares. Each row demonstrates the predicted scores of our model, effectively capturing the perceptual impact of varying distortions. 

The first row illustrates samples from the D2G dataset, where images with progressively increasing light intensity are shown. Our model’s predictions reflect an upward trend in quality scores as lighting improves, showcasing its sensitivity to perceptual clarity in low-light scenarios. However, excessive light can lead to distortions, as seen in the Flare7k examples (last two rows), where strong light flares reduce perceived quality. These examples validate DGIQA’s ability to identify degradation caused by unseen light scattering patterns and brightness distortions caused by lens flares.

Besides, the third row presents hazy images from the IHAZE dataset, a natural distortion type. Despite varying haze densities, our model demonstrates consistent quality predictions, indicating its ability to process and evaluate structural and textural details obscured by haze. This robustness showcases DGIQA's generalization performance and adaptability to diverse natural scene conditions.

Finally, the fourth and fifth rows feature motion blur distortions from the GoPro dataset. Here, DGIQA effectively distinguishes between varying degrees of blur, assigning lower scores to heavily blurred images while improving scores for less degraded samples. These results demonstrate our model’s capacity to evaluate dynamic distortions and maintain robust predictions even in challenging scenarios involving motion. Although the Koniq10k dataset used for training includes a handful of images with motion blur, the number of such samples are limited (less than $50$). Still, DGIQA is able to generalize well to the unseen levels of motion blurs of the GoPro dataset.

Overall, Figure~\ref{fig:final_prediction} highlights the adaptability of our model to diverse and complex natural distortions. By consistently providing perceptually aligned quality scores across varied datasets, DGIQA demonstrates its capability to generalize effectively and tackle real-world scenarios, further solidifying its robustness and versatility in NR-IQA tasks.

\end{document}